%% file: ArxivPreprint2027.tex
\documentclass[letterpaper]{article}
\usepackage[preprint]{aaai2027}
\usepackage[hyphens]{url}
\usepackage{graphicx}
\usepackage{natbib}
\usepackage{caption}
\usepackage{booktabs}
\usepackage{amsmath}
\usepackage{amssymb}
\usepackage{multirow}
\usepackage{colortbl}
\usepackage{algorithm}
\usepackage{algpseudocode}
\definecolor{modelrow}{RGB}{242,242,242}
\definecolor{rrprow}{RGB}{235,244,255}
\definecolor{avgcol}{RGB}{255,248,224}
\newcommand{\PhaseObservationWidth}{0.90\textwidth}

\input{math_commands}

\title{Beyond Global Routing Aggregation: Phase-Aware Expert Merging \\ for MoE Vision-Language Models}
\author{Hongyu Zhang, Cheng Yan, Xiang Xia, Wuyang Zhang\corresponding}
\affiliations{
University of Science and Technology of China, Hefei, China\\
\texttt{hongyuzhang@mail.ustc.edu.cn}, \texttt{wuyangz@ustc.edu.cn}
}

\begin{document}

\maketitle

\begin{abstract}
\input{sections/abstract}
\end{abstract}

\input{sections/introduction}
\input{sections/related_work}
\input{sections/preliminaries}
\input{sections/method}
\input{sections/main_results_table}
\input{sections/experiments}
\input{sections/conclusion}

{\small
\bibliography{references}
}

\clearpage
\appendix
\raggedbottom
\input{sections/appendix}

\end{document}

%% file: math_commands.tex
\newcommand{\RoleMerge}{\textsc{RoleMerge}}

%% file: sections/abstract.tex
Mixture-of-experts vision-language models (MoE-VLMs) increase model capacity with sparse expert activation, yet deployment requires storing the full expert pool.
Training-free expert merging reduces this burden, and many routing-based methods aggregate routing statistics across all tokens to determine merge compatibility.
However, MoE-VLM inference is phase-structured: image-context tokens carry visual content, question tokens specify the query, and answer tokens produce the output, with different counts and routing distributions.
Because image-context tokens are far more numerous, global aggregation can overemphasize image-context processing and obscure phase-conditioned expert roles, making experts serving different phases appear interchangeable and degrading model performance.
We therefore argue that MoE-VLM expert merging should preserve phase-conditioned expert roles, judging compatibility by how experts serve different phases rather than globally aggregated routing statistics.
Based on this view, we propose \RoleMerge{}, a training-free method that constructs each expert's Routing Role Profile (RRP) from phase-normalized routing statistics, capturing its relative phase preference.
Guided by expert--phase information loss, \RoleMerge{} merges experts with compatible profiles and their corresponding router entries while preserving answer-decoding expert distinctions.
Experiments on three models and multiple benchmarks show that \RoleMerge{} preserves more of the full model's performance than alternative expert-merging methods at matched expert-retention ratios, with relative improvements of up to 9.6\% in six-task macro-average performance.
These results validate phase-conditioned expert roles as a more effective basis than global routing aggregation for MoE-VLM expert merging.

%% file: sections/introduction.tex
\section{Introduction}

Mixture-of-experts (MoE) models increase model capacity by routing each token to only a small subset of experts, avoiding a proportional increase in per-token computation~\cite{shazeer2017outrageously,fedus2022switch}.
This conditional-computation paradigm has subsequently been incorporated into vision-language models (VLMs)~\cite{moma2024,i2moe2025}.
Mixture-of-experts vision-language models (MoE-VLMs) apply sparse expert activation to multimodal understanding and generation, increasing model capacity while keeping per-token computation manageable.
Recent representative MoE-VLMs include DeepSeek-VL2~\cite{deepseekvl22024}, Qwen3-VL~\cite{qwen3vl2025}, InternVL3.5~\cite{internvl35}, and Aria~\cite{li2024aria}, collectively validating this direction for scaling model capacity in VLMs.
However, sparse expert activation reduces per-token computation but not expert storage, because deployment must still store the full expert pool.

\input{Figures/fig_phase_observation}

This creates a practical need to reduce expert storage while preserving model performance.
Training-free expert merging directly addresses this need by replacing groups of pretrained experts with fewer merged experts.
Existing approaches address complementary parts of this process, including identifying and grouping merge-compatible experts~\cite{hcsmoe2025}, mitigating parameter conflicts~\cite{submoe2025,camex2025}, and determining how expert parameters within each group should be combined~\cite{puzzlemoe2025,namex2026}.
To decide which experts to merge, many routing-based methods aggregate routing statistics across all calibration tokens.
Together, they make training-free expert merging practical for reducing expert storage while largely preserving model performance.

However, directly applying this global aggregation to MoE-VLMs overlooks the distinct phase structure of multimodal inference.
A typical sequence comprises image-context tokens that encode visual content, question tokens that specify the query, and answer tokens that form the generated output.
These phases serve different functions, contribute substantially different numbers of tokens, and exhibit distinct routing distributions.
Consequently, globally aggregated routing statistics can be dominated by image-context processing, even though answer decoding may concentrate routing on a narrower subset of experts.
Figure~\ref{fig:phase_observation} shows that routing during answer decoding is more concentrated than during image-context processing across the analyzed models and benchmarks.
More fundamentally, global aggregation marginalizes out inference phase, so experts with different routing distributions across phases can nevertheless have identical globally aggregated routing statistics and appear merge-compatible.
Merging such experts may disrupt the routing structure needed for question processing or answer decoding and degrade model performance.
This raises the central question: \emph{how can we merge MoE-VLM experts to reduce storage while preserving the routing structure required across multimodal inference phases?}

Our key insight is to separate two decisions that global aggregation conflates: identifying how each expert serves different inference phases and determining whether two experts should be merged. We model the former as a phase-conditioned expert role and assess merge compatibility by the expert--phase information that would be lost through merging. Based on this reformulation, we propose \RoleMerge{}, a training-free expert-merging method that represents each expert's role with a Routing Role Profile (RRP), constructed from phase-normalized routing statistics to capture its relative preference across image-context processing, question processing, and answer decoding. Guided by expert--phase information loss, \RoleMerge{} groups experts with compatible profiles while preserving answer-decoding expert distinctions. After grouping, it combines the corresponding expert parameters and router entries using the same within-group weights, without retraining either the experts or the router.

\input{Figures/fig_method_overview}

Across three models, \RoleMerge{} retains more of the full model's macro-average performance than the comparison methods at all evaluated expert-retention ratios.
At $\rho=.50$, its relative macro-average gains over the best-performing comparison method reach 2.7\% on DeepSeek-VL2-Tiny and 9.6\% on Qwen3-VL-30B-A3B-Instruct.
On Qwen3-VL-30B-A3B-Instruct, it retains 85.5\% and 96.2\% of the full model's macro-average performance at $\rho=.50$ and $.75$, respectively.
Together, these results support our central premise that phase-conditioned expert roles provide a more effective basis than global routing aggregation for training-free MoE-VLM expert merging.

Our contributions are threefold:
\begin{itemize}
\item \textbf{Phase-conditioned expert roles.}
We reformulate MoE-VLM expert merging around phase-conditioned expert roles, characterizing how each expert serves different inference phases before assessing merge compatibility.
\item \textbf{Training-free expert merging.}
We propose \RoleMerge{}, which represents roles with RRPs, measures
compatibility by expert--phase information loss, preserves answer-decoding
distinctions, and merges expert and router parameters with shared
within-group weights.
\item \textbf{Empirical support for phase-conditioned expert roles.}
Across three models, six benchmarks, and three expert-retention ratios, \RoleMerge{} consistently retains more of the full model's macro-average performance than the comparison methods, with relative gains of up to 9.6\%.
Component ablations and calibration analyses further support the proposed reformulation.
\end{itemize}

%% file: Figures/fig_phase_observation.tex
\begin{figure*}[t]
  \centering
  \providecommand{\PhaseObservationWidth}{\textwidth}
  \includegraphics[width=\PhaseObservationWidth]{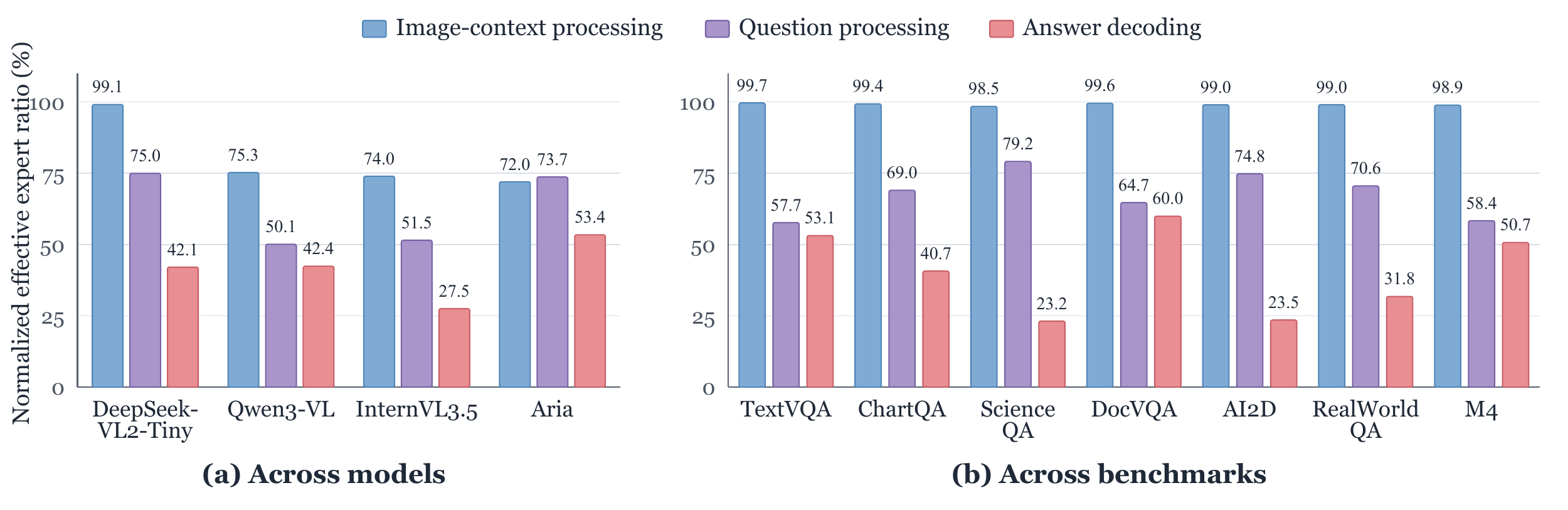}
  \caption{Phase-conditioned routing breadth, measured as exponentiated routing
entropy normalized by expert count.
Answer decoding uses fewer effective experts than image-context processing
across (a) four MoE-VLMs and (b) six QA datasets plus an external
M4-Instruct profile on DeepSeek-VL2-Tiny~\cite{li2024llavanextinterleave}.}
  \label{fig:phase_observation}
\end{figure*}

%% file: Figures/fig_method_overview.tex
\begin{figure*}[t]
  \centering
  \includegraphics[width=\textwidth]{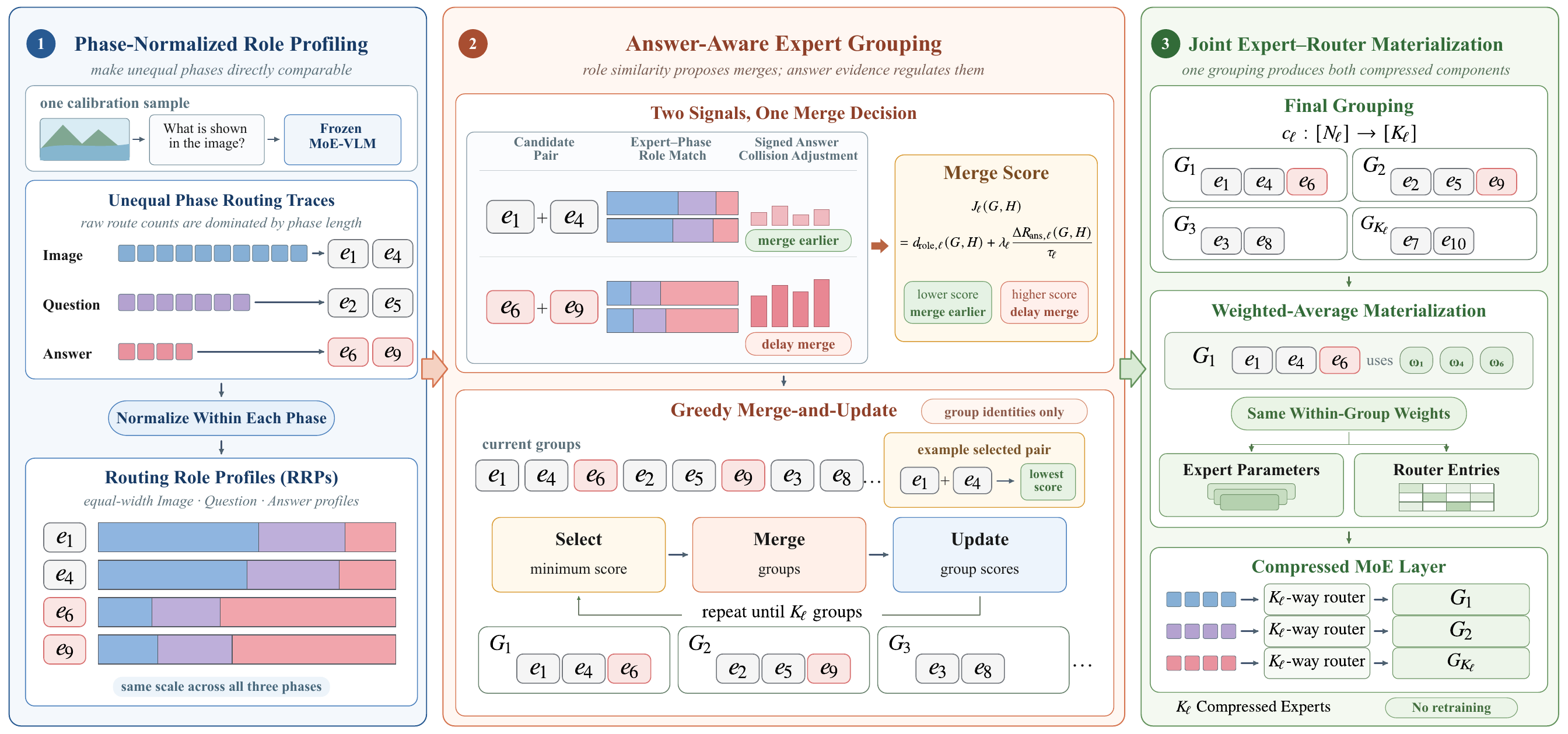}
  \caption{\RoleMerge{} overview.
Phase-separated routes form RRPs; answer-aware grouping combines role
dissimilarity and the signed collision adjustment, then merges expert
parameters and router entries with shared within-group weights.}
  \label{fig:method_overview}
\end{figure*}

%% file: sections/related_work.tex
\section{Related Work}

\paragraph{Sparse MoE-VLMs and deployment.}
Sparse MoEs increase model capacity by routing each token to only a small subset of experts, avoiding a proportional increase in per-token computation~\cite{shazeer2017outrageously,fedus2022switch}.
This conditional-computation paradigm has been extended to vision and multimodal modeling through V-MoE~\cite{vmoe2021}, MoMa~\cite{moma2024}, and I$^2$MoE~\cite{i2moe2025}, while recent MoE-VLMs include DeepSeek-VL2~\cite{deepseekvl22024}, Qwen3-VL~\cite{qwen3vl2025}, InternVL3.5~\cite{internvl35}, and Aria~\cite{li2024aria}.
However, sparse expert activation reduces per-token computation but not expert storage, because deployment must still store the full expert pool.
Distributed MoE systems improve execution efficiency through partitioning, communication, and scheduling~\cite{lepikhin2020gshard,rajbhandari2022deepspeedmoe,hwang2023tutel}, without reducing the expert pool.

\paragraph{Context-dependent routing and expert specialization.}
MoE routing varies with task, modality, token type, position, and language~\cite{vmoe2021,moma2024,i2moe2025,xue2024openmoe,muennighoff2025olmoe,bandarkar2026multilingualrouting}.
Generation-time traces also reveal concentrated expert reuse that can support caching and prefetching~\cite{xue2025moeinfinity}.
Together, these findings show that expert usage depends on token context rather than being fully characterized by a single global usage statistic.
This dependence is especially relevant to MoE-VLM inference, which comprises image-context tokens that carry visual content, question tokens that specify the query, and answer tokens that produce the output.
Prior work characterizes contextual specialization or exploits  routing for systems efficiency, but does not use phase-conditioned expert roles for merge compatibility.

\paragraph{Training-free expert merging.}
By replacing groups of pretrained experts with fewer merged experts, training-free expert merging directly reduces expert storage.
Existing methods explore behavior- or routing-based grouping~\cite{mcsmoe2024}, conflict-aware fusion~\cite{camex2025}, and structured or output-based merge realization~\cite{puzzlemoe2025,mergemoe2025}.
HC-SMoE groups experts through hierarchical clustering~\cite{hcsmoe2025}, whereas Sub-MoE mitigates parameter conflicts through subspace expert merging~\cite{submoe2025}.
NAMEx formulates expert fusion as a Nash-bargaining problem~\cite{namex2026}, while REAM combines router/output similarity, neuron alignment, and importance-weighted fusion~\cite{ream2026}.
These methods typically rely on parameter similarity, globally aggregated routing statistics, or calibration-set activations and outputs.
Routing-based criteria commonly aggregate routing statistics across all calibration tokens, marginalizing out inference phase and potentially making experts with different phase-conditioned roles appear merge-compatible.
\RoleMerge{} instead represents each expert with a Routing Role Profile constructed from phase-normalized routing statistics across image-context processing, question processing, and answer decoding, and uses these phase-conditioned expert roles to guide merge compatibility.

%% file: sections/preliminaries.tex
\section{Preliminaries and Problem Formulation}

\subsection{MoE Layers and Expert Merging}

For a positive integer $M$, let $[M]=\{1,\ldots,M\}$.
Consider an MoE-VLM with $L$ MoE layers.
For each layer $\ell\in[L]$, let $N_\ell$ denote the number of experts in its expert pool, $\phi_{\ell,e}$ the function implemented by expert $e$, and $R_\ell$ the router.
Given a token representation $\mathbf{x}\in\mathbb{R}^{d}$, the router selects $k$ experts and assigns them normalized routing weights $g_{\ell,e}(\mathbf{x})$.
The resulting output of MoE layer $\ell$ is
\begin{equation}
\mathbf{y}_\ell(\mathbf{x})
=
\sum_{e\in\operatorname{TopK}(R_\ell(\mathbf{x}),k)}
g_{\ell,e}(\mathbf{x})\phi_{\ell,e}(\mathbf{x}).
\label{eq:moe_layer}
\end{equation}
We use $\Theta_{\ell,e}$ to denote the parameters of expert $e$ and $\mathbf{r}_{\ell,e}$ to denote its corresponding expert-indexed output parameter in the pretrained router prior to expert merging.

Given a target expert-retention ratio $\rho$, expert merging reduces layer $\ell$ to $K_\ell$ expert groups, where $K_\ell/N_\ell=\rho$ for the evaluated ratios.
A surjective grouping map $c_\ell:[N_\ell]\rightarrow[K_\ell]$ induces the partition
$\mathcal{G}_\ell=\{c_\ell^{-1}(k):k\in[K_\ell]\}$.
The problem is to group experts while preserving the routing structure required for multimodal inference.

\subsection{Limitation of Global Routing Aggregation}

Let $\mathcal{B}=\{\mathrm{img},\mathrm{ques},\mathrm{ans}\}$ denote image-context processing, question processing, and answer decoding.
For a calibration set $\mathcal{D}_{\mathrm{cal}}$, let
$n_{\ell,b}(e)$ be the number of times expert $e$ is selected by the
router at layer $\ell$ for tokens processed during phase $b$, and let
$T_{\ell,b}=\sum_e n_{\ell,b}(e)$ denote the total number of expert
selections in that phase.
The phase-normalized routing distribution is
$p_{\ell,b}(e)=n_{\ell,b}(e)/T_{\ell,b}$, and the empirical fraction
of selections contributed by phase $b$ is
$\widehat{\pi}_\ell(b)=T_{\ell,b}/\sum_{b'}T_{\ell,b'}$.
The resulting global routing distribution is
\begin{equation}
\overline{p}_\ell(e)
=
\sum_{b\in\mathcal{B}}
\widehat{\pi}_\ell(b)p_{\ell,b}(e).
\label{eq:global_routing_aggregation}
\end{equation}
Because image-context processing typically contains substantially more
tokens, it contributes more expert selections and therefore receives
greater weight in $\overline{p}_\ell(e)$, potentially obscuring
differences expressed during question processing or answer decoding in multimodal inference.

Global aggregation also marginalizes inference phase: distinct
$\{p_{\ell,b}(e)\}_{b\in\mathcal{B}}$ can yield the same
$\overline{p}_\ell(e)$.
Experts with different phase roles may therefore appear merge-compatible.
Appendix~A constructs such a non-identifiability example.

\subsection{Phase-Conditioned Expert Roles}

We instead characterize how each expert serves the three inference phases before assessing merge compatibility.
To prevent unequal token counts from dominating this characterization, we use the uniform phase prior
$\pi(b)=1/|\mathcal{B}|$ and construct
$P_\ell(e,b)=\pi(b)p_{\ell,b}(e)$, with
$P_\ell(e)=\sum_b P_\ell(e,b)$.

Let $S_\ell\in[N_\ell]$ denote expert identity and $B\in\mathcal{B}$ denote inference phase under this constructed joint distribution.
For $P_\ell(e)>0$, the Routing Role Profile (RRP) of expert $e$ is
\begin{equation}
q_{\ell,e}(b)
=
P_\ell(B=b\mid S_\ell=e)
=
\frac{P_\ell(e,b)}{P_\ell(e)}.
\label{eq:rrp}
\end{equation}
For an expert with $P_\ell(e)=0$, we use the zero vector by convention.
The RRP represents the expert's phase-conditioned expert role by describing its relative preference across image-context processing, question processing, and answer decoding.
It is induced by routing behavior on the calibration set rather than interpreted as a universal semantic label.

This formulation separates two decisions coupled by global routing aggregation: identifying how an expert serves different inference phases and determining whether two experts should be merged.
The RRP characterizes the former, while the expert marginal $P_\ell(e)$ weights the information lost when evaluating a candidate merge.

\subsection{Expert--Phase Information Loss}

We quantify merge compatibility by the expert--phase information removed by grouping.
Let $C_\ell=c_\ell(S_\ell)$ denote the retained group identity.
For a group $G\in\mathcal{G}_\ell$, let
$P_\ell(G)=\sum_{e\in G}P_\ell(e)$ and, when $P_\ell(G)>0$,
$q_{\ell,G}=\sum_{e\in G}P_\ell(e)q_{\ell,e}/P_\ell(G)$.
For a zero-mass group, we use the zero vector, and terms weighted by $P_\ell(e)=0$ are defined as zero.
The expert--phase information loss of the grouping is
\begin{align}
\Delta_{\mathrm{EP},\ell}(c_\ell)
&=
I_{P_\ell}(S_\ell;B)-I_{P_\ell}(C_\ell;B)
\nonumber\\
&=
\sum_{G\in\mathcal{G}_\ell}\sum_{e\in G}
P_\ell(e)
D_{\mathrm{KL}}
\left(q_{\ell,e}\Vert q_{\ell,G}\right),
\label{eq:group_ep_information_loss}
\end{align}
where mutual information is evaluated under $P_\ell$ and
$D_{\mathrm{KL}}$ denotes KL divergence over $\mathcal{B}$.

For two experts $i\neq j$ with $P_\ell(i)+P_\ell(j)>0$, let
$m_{\ell,ij}=
\bigl(P_\ell(i)q_{\ell,i}+P_\ell(j)q_{\ell,j}\bigr)/
\bigl(P_\ell(i)+P_\ell(j)\bigr)$.
Their pairwise expert--phase information loss is
\begin{align}
d_{\mathrm{EP},\ell}(i,j)
={}&
P_\ell(i)
D_{\mathrm{KL}}
\left(q_{\ell,i}\Vert m_{\ell,ij}\right)
\nonumber\\
&+
P_\ell(j)
D_{\mathrm{KL}}
\left(q_{\ell,j}\Vert m_{\ell,ij}\right).
\label{eq:expert_phase_information_loss}
\end{align}
This quantity is the exact decrease in expert--phase mutual information
caused by merging $i$ and $j$ while leaving all other experts as
one-expert groups.
Appendix~A provides the corresponding derivation and its relation to
the group-level expert--phase information loss.
When $P_\ell(i)+P_\ell(j)=0$, we define
$d_{\mathrm{EP},\ell}(i,j)=0$.
A small value indicates compatible phase-conditioned expert roles.

Expert--phase information loss distinguishes how experts serve inference phases.
However, it alone cannot prevent co-grouping of experts frequently selected in
the same phase, especially during answer decoding.
We therefore augment it to preserve answer-decoding expert distinctions.

%% file: sections/method.tex
\section{\RoleMerge}
\subsection{Pipeline}
Given a pretrained MoE-VLM, a calibration set
$\mathcal{D}_{\mathrm{cal}}$, and a target expert-retention ratio
$\rho$, \RoleMerge{} reduces the expert pool of each selected MoE layer
$\ell$ from $N_\ell$ to $K_\ell$ experts, where
$K_\ell/N_\ell=\rho$. It processes each layer in four steps. First, it
collects routing statistics separately for the three inference phases:
image-context processing, question processing, and answer decoding.
Second, it constructs Routing Role Profiles (RRPs) that represent the
phase-conditioned expert roles and computes the pairwise expert--phase
information losses. Third, it greedily merges expert groups until
$K_\ell$ groups remain, using an answer-aware merge score that accounts
for role compatibility and preserves answer-decoding expert
distinctions. Finally, it merges the expert and router
parameters using the same within-group weights, producing $K_\ell$
merged experts and a router with a reduced output dimension. 
The following subsections detail RRP construction, the answer-aware
merge score, and expert--router merging; Appendix~B provides
implementation details for exact reproduction.

\subsection{Routing Role Profile Construction}
For each selected layer $\ell$, \RoleMerge{} collects the
phase-separated routing counts $n_{\ell,b}(e)$ over
$\mathcal D_{\mathrm{cal}}$. Using the phase-normalized distributions,
uniform phase prior, and RRP construction defined in the preliminaries,
it obtains $q_{\ell,e}$ and the pairwise expert--phase information
losses $d_{\mathrm{EP},\ell}(i,j)$. The answer-decoding distribution
$p_{\ell,\mathrm{ans}}(e)$ is retained for the signed collision adjustment in the
next step.

\subsection{Answer-Aware Merge Score}
Given $d_{\mathrm{EP},\ell}(i,j)$ and $p_{\ell,\mathrm{ans}}(e)$,
\RoleMerge{} scores each candidate group merge using role dissimilarity
and a signed adjustment based on answer-decoding mass.

We first normalize $d_{\mathrm{EP},\ell}(i,j)$ within each layer and define the group-wise role dissimilarity for candidate groups $G,H\subseteq[N_\ell]$ using the following single-link expression:
\begin{equation}
d_{\mathrm{role},\ell}(G,H)
=
\min_{i\in G,j\in H}
\frac{d_{\mathrm{EP},\ell}(i,j)}
{\max_{\substack{u,v\in[N_\ell]\\u\ne v}}
d_{\mathrm{EP},\ell}(u,v)}.
\end{equation}
If the layer-wise maximum in the denominator is zero, we set
$d_{\mathrm{role},\ell}(G,H)=0$ for all candidate group pairs.
This definition applies single linkage: two groups have a small
$d_{\mathrm{role},\ell}(G,H)$ when they contain at least one expert
pair with compatible profiles, as indicated by low expert--phase
information loss. It is a greedy extension of the exact pairwise losses
that propagates local profile compatibility during group construction.
For multi-expert groups, $d_{\mathrm{role},\ell}(G,H)$ is not generally the group-level loss in Eq.~\eqref{eq:group_ep_information_loss} and does not guarantee a globally optimal grouping.

To preserve answer-decoding expert distinctions beyond role
compatibility, we construct a signed collision adjustment from
$p_{\ell,\mathrm{ans}}(e)$. We measure the answer-decoding routing
concentration of layer $\ell$ by
$\kappa_\ell=1-H(p_{\ell,\mathrm{ans}})/\log N_\ell$, where
$H(\cdot)$ denotes Shannon entropy. 
We set $\lambda_\ell\propto\max(\kappa_\ell,\varepsilon)$ and normalize it to mean $\lambda$ over the selected layers, with $\varepsilon>0$ providing a positive floor for each layer.

For a candidate group $G$, the ratio between its total answer-decoding
routing mass and the largest individual mass within the group indicates
whether that mass is concentrated on one expert or distributed across
multiple experts. The ratio is close to one when a single expert
dominates and increases when several experts carry substantial
answer-decoding routing mass. We therefore use the group-level collision measure:
\begin{equation}
R_{\mathrm{ans},\ell}(G)
=
\left[
\frac{\sum_{e\in G}p_{\ell,\mathrm{ans}}(e)}
{\max_{e\in G}p_{\ell,\mathrm{ans}}(e)+\varepsilon}
-1
\right]_{+}.
\end{equation}
For a candidate merge, its signed collision adjustment is
$\Delta R_{\mathrm{ans},\ell}(G,H)
=R_{\mathrm{ans},\ell}(G\cup H)
-R_{\mathrm{ans},\ell}(G)
-R_{\mathrm{ans},\ell}(H)$.
Because $R_{\mathrm{ans},\ell}$ is group-normalized and nonadditive,
this adjustment can be either positive or negative. A positive value
raises the merge score and lowers the priority of the candidate merge,
whereas a negative value lowers the score and raises its priority.
A negative value indicates that the union carries less normalized
collision than the two groups considered separately, so the merge does
not introduce additional answer-expert overlap under this score.

To normalize the magnitude of the collision adjustment across layers,
let $\tau_\ell>0$ be the median of the strictly positive collision
increments computed over original expert pairs, with a fallback value
of $1$ when no such increment exists.
The resulting score for merging groups $G$ and $H$ is
\begin{equation}
J_{\ell}(G,H)
=
d_{\mathrm{role},\ell}(G,H)
+
\lambda_{\ell}
\frac{\Delta R_{\mathrm{ans},\ell}(G,H)}
{\tau_{\ell}}.
\label{eq:grouping}
\end{equation}
At each iteration, \RoleMerge{} merges the candidate pair with the
smallest $J_\ell(G,H)$ until $K_\ell$ groups remain, producing the grouping map $c_\ell$ used for parameter merging.

Unlike the layer-aggregated answer-decoding distribution used in the
merge score, ACR is computed separately for each task--layer calibration
record $r$. Given its normalized answer-decoding routing distribution
$p^A_r(e)$, we define
Answer-Carrier Recoverability as
$\operatorname{ACR}_r(c_\ell)
=\sum_{G\in\mathcal G_\ell}\max_{e\in G}p^A_r(e)$.
ACR diagnoses how well $c_\ell$ retains distinctions among experts used
during answer decoding; it is not optimized by $J_\ell$. Its derivation
is provided in Appendix~A. The grouping map $c_\ell$ then determines the merging of expert and router parameters.

\subsection{Merging Expert and Router Parameters}
Given the grouping map $c_\ell$ produced by the preceding step,
\RoleMerge{} merges the parameters in each group
$G\in\mathcal G_\ell$ to form one merged expert and its corresponding
router entry.

Following prior frequency-weighted expert merging~\cite{mcsmoe2024},
we define $\omega_{\ell,e}$ by normalizing within group $G$ the original
calibration routing count pooled across inference phases,
$\sum_{b\in\mathcal B}n_{\ell,b}(e)$, such that
$\sum_{e\in G}\omega_{\ell,e}=1$. 
Applying the shared within-group weights to both parameter sets gives
\begin{equation}
\widehat{\Theta}_{\ell,G}
=
\sum_{e\in G}\omega_{\ell,e}\Theta_{\ell,e},
\qquad
\widehat{\mathbf r}_{\ell,G}
=
\sum_{e\in G}\omega_{\ell,e}\mathbf r_{\ell,e}.
\end{equation}

This yields $K_\ell$ merged experts and reduces the router output to $K_\ell$.
Only routed expert parameters and router output entries change; the
visual encoder, attention modules, shared parameters, and decoding
interface remain fixed.

%% file: sections/main_results_table.tex
\begin{table*}[t]
\centering
{
\small
\setlength{\tabcolsep}{0.40mm}
\begin{tabular}{@{}c*{6}{c}>{\columncolor{avgcol}}c@{\hspace{1.2mm}}*{6}{c}>{\columncolor{avgcol}}c@{\hspace{1.2mm}}*{6}{c}>{\columncolor{avgcol}}c@{}}
\toprule
\multirow{2}{*}{\textbf{Method}}
& \multicolumn{7}{c}{\textbf{$\rho=.50$}}
& \multicolumn{7}{c}{\textbf{$\rho=.625$}}
& \multicolumn{7}{c}{\textbf{$\rho=.75$}} \\
\cmidrule(lr){2-8}\cmidrule(lr){9-15}\cmidrule(l){16-22}
& Info & MMU & OCR & ARC & PIQA & Wino & \textbf{Avg}
& Info & MMU & OCR & ARC & PIQA & Wino & \textbf{Avg}
& Info & MMU & OCR & ARC & PIQA & Wino & \textbf{Avg} \\
\midrule
\rowcolor{modelrow}
\multicolumn{22}{@{}c@{}}{\textbf{DeepSeek-VL2-Tiny}} \\
Full
& 49.3 & 39.1 & 81.1 & 67.9 & 73.2 & 59.9 & 61.7
& 49.3 & 39.1 & 81.1 & 67.9 & 73.2 & 59.9 & 61.7
& 49.3 & 39.1 & 81.1 & 67.9 & 73.2 & 59.9 & 61.7 \\
MC-SMoE
& 31.7 & 23.8 & 48.6 & 34.2 & 56.7 & 45.9 & 40.2 
& 39.6 & 26.2 & 60.4 & 37.7 & 60.5 & 46.2 & 45.1
& 45.3 & 31.3 & 68.5 & 42.6 & 62.4 & 49.7 & 50.0 \\
MergeMoE
& 17.4 & 20.3 & 12.6 & 30.9 & 55.9 & 45.2 & 30.4 
& 30.9 & 25.4 & 43.5 & 43.8 & 64.5 & 48.2 & 42.7
& 41.8 & 33.8 & 70.2 & 58.5 & 69.8 & 55.1 & 54.9 \\
HC-SMoE
& 18.5 & 25.8 & 49.3 & 45.2 & 63.1 & 51.9 & 42.3
& 31.1 & 27.9 & 60.5 & 52.1 & 64.4 & 53.8 & 48.3
& 39.2 & 32.3 & 68.7 & 57.6 & 69.2 & 54.8 & 53.6 \\
REAM
& \underline{40.3} & \textbf{34.1} & \underline{70.2} & 46.4 & \textbf{67.3} & 53.0 & \underline{51.9}
& \underline{44.0} & \underline{34.8} & \underline{72.9} & 51.8 & \underline{68.8} & \textbf{57.4} & \underline{54.9}
& 46.5 & \underline{35.4} & \textbf{78.2} & 54.0 & 69.0 & \underline{57.9} & 56.8 \\
Sub-MoE
& 28.3 & 27.6 & 42.3 & 48.2 & 62.0 & \textbf{55.0} & 43.9
& 34.8 & 30.4 & 59.4 & 53.9 & 65.5 & \underline{57.2} & 50.2
& 43.7 & 35.0 & 72.5 & 55.7 & 66.9 & 55.6 & 54.9 \\
NAMEx
& 30.6 & 25.9 & 61.3 & \textbf{51.9} & 63.2 & \underline{53.9} & 47.8
& 37.3 & 30.3 & 64.3 & \underline{57.7} & 67.0 & 54.1 & 51.8
& \underline{48.1} & 34.7 & 71.1 & \underline{60.8} & \underline{70.2} & \textbf{58.6} & \underline{57.2} \\
\rowcolor{rrprow}
\RoleMerge
& \textbf{49.3} & \underline{32.3} & \textbf{70.4} & \underline{49.9} & \underline{65.1} & 52.7 & \textbf{53.3}
& \textbf{50.7} & \textbf{35.1} & \textbf{74.2} & \textbf{58.9} & \textbf{69.4} & 53.3 & \textbf{56.9}
& \textbf{59.0} & \textbf{37.1} & \underline{77.6} & \textbf{62.2} & \textbf{70.4} & 56.1 & \textbf{60.4} \\
\midrule
\rowcolor{modelrow}
\multicolumn{22}{@{}c@{}}{\textbf{DeepSeek-VL2-Small}} \\
Full
& 57.9 & 46.0 & 81.3 & 77.9 & 77.5 & 70.2 & 68.5
& 57.9 & 46.0 & 81.3 & 77.9 & 77.5 & 70.2 & 68.5
& 57.9 & 46.0 & 81.3 & 77.9 & 77.5 & 70.2 & 68.5 \\
Sub-MoE
& 45.3 & \underline{34.4} & \underline{64.9} & \underline{35.9} & 59.0 & \textbf{51.9} & \underline{48.6}
& \underline{48.2} & \underline{37.1} & \underline{71.5} & \underline{40.4} & 61.7 & \underline{54.3} & \underline{52.2}
& \textbf{58.3} & \underline{41.0} & \underline{74.8} & \underline{46.1} & \underline{65.8} & \underline{55.9} & \underline{57.0} \\
NAMEx
& \underline{45.8} & 34.2 & 63.8 & 35.5 & \underline{59.1} & \underline{51.4} & 48.3
& 48.1 & 36.8 & 71.3 & 39.5 & \underline{62.1} & \textbf{54.5} & 52.0
& \underline{57.7} & \underline{41.0} & \underline{74.8} & 45.3 & 65.5 & \underline{55.9} & 56.7 \\
\rowcolor{rrprow}
\RoleMerge
& \textbf{51.5} & \textbf{39.0} & \textbf{69.8} & \textbf{36.2} & \textbf{62.2} & 49.9 & \textbf{51.4}
& \textbf{54.5} & \textbf{42.0} & \textbf{74.9} & \textbf{43.4} & \textbf{66.9} & 52.5 & \textbf{55.7}
& \textbf{58.3} & \textbf{44.8} & \textbf{78.3} & \textbf{48.7} & \textbf{68.5} & \textbf{56.4} & \textbf{59.2} \\
\midrule
\rowcolor{modelrow}
\multicolumn{22}{@{}c@{}}{\textbf{Qwen3-VL-30B-A3B-Instruct}} \\
Full
& 80.1 & 54.4 & 84.9 & 79.1 & 79.3 & 73.8 & 75.3
& 80.1 & 54.4 & 84.9 & 79.1 & 79.3 & 73.8 & 75.3
& 80.1 & 54.4 & 84.9 & 79.1 & 79.3 & 73.8 & 75.3 \\
Sub-MoE
& 57.0 & 37.4 & 65.6 & \underline{61.8} & 68.6 & \textbf{59.7} & 58.4
& 64.5 & 42.9 & 74.0 & \underline{68.4} & 73.4 & \underline{64.0} & 64.5
& 71.3 & 48.7 & 78.9 & 73.7 & 76.3 & \underline{68.8} & 69.6 \\
NAMEx
& \underline{57.8} & \underline{38.1} & \underline{67.9} & 60.8 & \underline{69.3} & \underline{58.4} & \underline{58.7}
& \underline{64.9} & \underline{43.2} & \underline{74.4} & 68.3 & \underline{75.6} & \textbf{65.3} & \underline{65.3}
& \underline{73.3} & \underline{49.7} & \underline{80.6} & \underline{73.8} & \textbf{78.6} & \textbf{69.9} & \underline{71.0} \\
\rowcolor{rrprow}
\RoleMerge
& \textbf{72.5} & \textbf{44.6} & \textbf{74.4} & \textbf{66.9} & \textbf{71.1} & 56.5 & \textbf{64.3}
& \textbf{76.3} & \textbf{47.7} & \textbf{78.2} & \textbf{73.9} & \textbf{76.5} & 62.7 & \textbf{69.2}
& \textbf{79.7} & \textbf{52.3} & \textbf{81.2} & \textbf{76.1} & \underline{77.6} & 67.6 & \textbf{72.4} \\
\bottomrule
\end{tabular}
}
\caption{Main compression results on official 0--100 scores (higher is better).
InfoVQA uses ANLS$\times100$; all others use accuracy$\times100$.
$\rho$: expert retention; TCS: TextVQA--ChartQA--ScienceQA calibration;
Avg: unweighted six-task mean before rounding.
Shading marks Avg; bold/underline mark the best/second-best compressed values per column.}
\label{tab:main_results}
\end{table*}

%% file: sections/experiments.tex
\section{Experiments}
\label{sec:experiments}

We organize the experiments around four questions that match the main claims and boundaries of \RoleMerge{}.
\begin{enumerate}
\item[\textbf{Q1}] Does \RoleMerge{} retain more of the full model's performance than comparison methods under matched expert-retention ratios and calibration settings?
\item[\textbf{Q2}] How do phase-conditioned expert roles and the signed collision adjustment contribute to \RoleMerge{}?
\item[\textbf{Q3}] How sensitive is \RoleMerge{} to calibration construction, the collision weight, and the grouping rule?
\item[\textbf{Q4}] How much does \RoleMerge{} reduce expert storage and improve inference efficiency in practice?
\end{enumerate}

\subsection{Experimental Settings}
To evaluate the effectiveness of \RoleMerge{} for MoE-VLM compression, we conduct experiments on three models: DeepSeek-VL2-Tiny and DeepSeek-VL2-Small~\cite{deepseekvl22024}, together with Qwen3-VL-30B-A3B-Instruct~\cite{qwen3vl2025}.
We use the uncompressed Full Model as the reference and compare \RoleMerge{} with Sub-MoE~\cite{submoe2025} and NAMEx~\cite{namex2026} across all three models.
On DeepSeek-VL2-Tiny, we additionally include MC-SMoE~\cite{mcsmoe2024}, MergeMoE~\cite{mergemoe2025}, HC-SMoE~\cite{hcsmoe2025}, and REAM~\cite{ream2026}.

\paragraph{Models and compression protocol.}
The main compression points are $\rho\in\{.50,.625,.75\}$.
At fixed model and $\rho$, all methods share the same checkpoint,
calibration set, retained expert count, and compressed MoE layers.
The default calibration set is TextVQA~\cite{singh2019textvqa}--ChartQA~\cite{masry2022chartqa}--ScienceQA~\cite{lu2022scienceqa} (TCS), selected before downstream evaluation using calibration-only routing diagnostics.
The calibration procedure uses only routing traces and does not use task labels, answer correctness, or downstream scores.
Using single linkage, \RoleMerge{} extends normalized pairwise
expert--phase losses to group dissimilarities, adds the signed collision
adjustment ($\lambda=.03$), and merges expert and router parameters with the same within-group weights during materialization.

\paragraph{Benchmarks and metrics.}
We evaluate all methods on six benchmarks: InfoVQA~\cite{infovqa2022}, MMMU~\cite{mmmu2024}, and OCRBench~\cite{ocrbench2023} for multimodal understanding, and ARC-Easy~\cite{arc2018}, PIQA~\cite{piqa2020}, and Winogrande~\cite{winogrande2020} for text reasoning.
We map the official metrics to a common 0--100 scale, where higher is better: InfoVQA uses ANLS$\times100$, and the other five benchmarks use official accuracy$\times100$.
With a fixed calibration set, \RoleMerge{} is deterministic; all autoregressive generation uses greedy decoding with temperature set to 0 and sampling disabled, and methods share fixed evaluation examples and order.
MM3 averages the three multimodal task scores, Text3 averages the three text task scores, and Avg is the unweighted average over all six.
Full Model denotes the uncompressed reference, and bold numbers mark the best compressed method under the same model, retention ratio, and task column.
Unless otherwise stated, experiments use two NVIDIA A100 80GB GPUs for the largest model; the efficiency audit uses one A100 80GB.
Appendix~C details the experimental protocol.

\subsection{Main Results under Matched Compression}

\textbf{To answer Q1.} Table~\ref{tab:main_results} presents matched comparisons across three MoE-VLM settings and three retention ratios using TCS calibration under matched within-model settings.
On DeepSeek-VL2-Tiny, \RoleMerge{} achieves the best compressed average
at all three retention ratios, retaining 86.3\%, 92.2\%, and 97.8\% of
Full Model Avg at $\rho=.50$, $.625$, and $.75$, respectively.
At $\rho=.75$, \RoleMerge{} reaches 60.4 Avg, compared with the best competing Avg of 57.2, a 5.6\% relative gain.
When the parameter scale increases to DeepSeek-VL2-Small, the same pattern remains: \RoleMerge{} again gives the best Avg at all three retention ratios and retains 75.1\%, 81.3\%, and 86.4\% of the Full Model Avg.
At $\rho=.625$, it reaches 55.7 Avg, improving by 6.7\% over the strongest compared method, Sub-MoE (52.2), while also leading on the three multimodal tasks.
On Qwen3-VL-30B-A3B-Instruct, the result transfers to a different MoE-VLM family and architecture.
\RoleMerge{} retains 85.5\%, 91.9\%, and 96.2\% of the Full Model Avg across the three retention ratios; at $\rho=.50$, it improves over NAMEx from 58.7 to 64.3 Avg, a 9.6\% relative gain.
The relative margins narrow as more experts are kept, but the best compressed Avg remains \RoleMerge{} in all nine model--retention settings.
Appendix~D reports paired uncertainty estimates and additional Q1 comparisons across models and retention ratios.

\subsection{Mechanism Analysis}

\input{sections/analysis_tables}

\textbf{To answer Q2.}
Table~\ref{tab:ablation_summary} reports matched mechanism controls on DeepSeek-VL2-Small at $\rho=.625$.
All controls use the same calibration set, retained expert count, grouping rule, and parameter-merging procedure, with only the specified mechanism changed.
ACR measures preservation of answer-decoding expert distinctions and
serves as a mechanism diagnostic, not a universal performance predictor.

Pooling routing statistics from image-context processing, question processing, and answer decoding lowers ACR from 95.0\% to 93.1\% ($-1.9$ points) and MM3 from 57.1 to 53.5 ($-3.6$), showing the benefit of retaining phase identity.
Permuting RRPs across expert identities produces a larger degradation: ACR falls to 70.8\% ($-24.3$) and MM3 to 51.7 ($-5.5$), supporting the importance of preserving the correspondence between each expert and its RRP.
This result shows that phase separation alone is insufficient: each RRP must remain associated with the expert whose phase-conditioned role it represents.
Together, these controls support phase-conditioned expert roles for assessing merge compatibility.

Removing the signed collision adjustment reduces ACR to 89.1\% ($-5.9$) and MM3 to 55.4 ($-1.7$).
The simultaneous decreases support the intended role of the adjustment in preserving answer-decoding expert distinctions, complementing the role-compatibility signal from expert--phase information loss.
Appendix~E gives expanded controls; Appendix~A reports an additional signed-collision diagnostic.

\subsection{Calibration and Sensitivity Analysis}

\textbf{To answer Q3.} We evaluate sensitivity to calibration construction, the collision weight, and the grouping rule.

\paragraph{Calibration composition.}
Under a fixed calibration budget, the one-source, two-source, and TCS calibration sets yield similar two-task macro scores of 39.0, 39.8, and 39.7 points, respectively.
This result suggests that the TCS calibration set is not unusually sensitive to the number of calibration sources.
Alternative equal-prior mixtures can score higher on the two-task diagnostic (DAR reaches 42.1 points), so we do not claim TCS is uniquely optimal.
Instead, TCS is kept as a pre-specified, balanced default because it has the lowest top-16 answer-decoding expert overlap (.442) and highest union coverage (.411) among the audited mixtures.

\paragraph{Calibration source.}
On the Tiny three-retention-ratio sweep, a mixed multimodal calibration set reaches 57.6 points, compared with 51.2 for text-only answer traces and 53.1 for a task-specific multimodal source.
This gap supports constructing RRPs from multimodal calibration inputs.
TCS averages 56.9, close to 56.8 for an alternative multimodal calibration set, suggesting stable performance across different multimodal calibration constructions.

\paragraph{Calibration-set size.}
The calibration-size sweep ranges from 54.9 to 57.6 over 32--1024 examples per source, indicating a broad plateau rather than sharp size sensitivity.

\paragraph{Collision weight and grouping rule.}
Table~\ref{tab:sensitivity_summary} shows that the default setting is not a sharp optimum.
Increasing $\lambda$ from .03 to .10 changes the six-task macro by $+0.6$ points, whereas setting $\lambda=0$ drops the macro by 2.6 points.
Among the audited grouping rules, single link gives the strongest downstream preservation.
In particular, $k$-means++ trails single link by 5.4 points, indicating that centroid-based clustering is a poor default in this setting.
Appendix~F provides the full calibration and sensitivity analyses.

\subsection{Storage and Efficiency Analysis}

\textbf{To answer Q4.} For Qwen3-VL-30B-A3B-Instruct at $\rho=.625$, \RoleMerge{} reduces routed-expert parameters by 37.5\% (28.99B to 18.12B) and total parameters and serialized bf16 storage by 35.0\% (31.07B to 20.19B; 62.14GB to 40.39GB).
In a controlled single-A100 audit of DeepSeek-VL2-Small at $\rho=.50$, it reduces steady-state and peak incremental GPU memory by 33.8\% and 32.2\%, TTFT by 32.1\%, and end-to-end latency by 15.3\%, while increasing 32-token throughput by 5.7\%.
All checkpoints remain GPU-resident, and these directions are consistent across repetitions.
Appendix~G details the efficiency and storage measurements.

%% file: sections/analysis_tables.tex
\begin{table}[t]
\centering
\small
\setlength{\tabcolsep}{2.4pt}
\begin{tabular}{@{}lrrrr@{}}
\toprule
Variant & ACR (\%) & $\Delta$ACR & MM3 & $\Delta$MM3 \\
\midrule
w/o phase split & 93.1 & $-1.9$ & 53.5 & $-3.6$ \\
w/o collision adj. & 89.1 & $-5.9$ & 55.4 & $-1.7$ \\
permuted roles & 70.8 & $-24.3$ & 51.7 & $-5.5$ \\
\rowcolor{rrprow}
Complete \RoleMerge{} & \textbf{95.0} & 0.0 & \textbf{57.1} & 0.0 \\
\bottomrule
\end{tabular}
\caption{DeepSeek-VL2-Small component ablations at $\rho=.625$. ACR averages 78 task--layer records. Deltas relative to the complete \RoleMerge{} setting are percentage points for ACR and normalized points for MM3, the mean of multimodal benchmarks.}
\label{tab:ablation_summary}
\end{table}

\begin{table}[t]
\centering
\small
\setlength{\tabcolsep}{1.4pt}
\begin{tabular}{@{}lccccc@{}}
\toprule
Factor & Setting & MM3 & Text3 & Avg & $\Delta$ \\
\midrule
& $\lambda=0$ & 50.1 & 58.7 & 54.4 & $-2.6$ \\
& $\lambda=.01$ & 53.9 & 59.3 & 56.6 & $-0.4$ \\
& \cellcolor{rrprow}$\lambda=.03$ (ours)
& \cellcolor{rrprow}53.4
& \cellcolor{rrprow}60.5
& \cellcolor{rrprow}56.9
& \cellcolor{rrprow}0.0 \\
\multirow{-4}{*}{\shortstack[l]{\textit{Collision}\\\textit{weight}}}
& $\lambda=.10$ & \textbf{53.9} & \textbf{61.2} & \textbf{57.6} & 0.6 \\
\midrule
& avg. link & 52.9 & 58.8 & 55.9 & $-1.1$ \\
& comp. link & 52.3 & 58.3 & 55.3 & $-1.6$ \\
& $k$-means++ & $43.7_{\scriptscriptstyle 2.3}$ & $59.2_{\scriptscriptstyle 0.9}$ & $51.5_{\scriptscriptstyle 1.2}$ & $-5.4$ \\
\multirow{-4}{*}{\shortstack[l]{\textit{Grouping}\\\textit{rule}}}
& \cellcolor{rrprow}single link (ours)
& \cellcolor{rrprow}\textbf{53.4}
& \cellcolor{rrprow}\textbf{60.5}
& \cellcolor{rrprow}\textbf{56.9}
& \cellcolor{rrprow}0.0 \\
\bottomrule
\end{tabular}
\caption{DeepSeek-VL2-Tiny sensitivity at $\rho=.625$
(normalized points; higher is better).
$\Delta$ is the Avg change from $\lambda=.03$/single link;
The $k$-means++ row gives five-seed means; subscripts denote standard deviations.}
\label{tab:sensitivity_summary}
\end{table}

%% file: sections/conclusion.tex
\section{Conclusion and Discussion}

In this paper, we reformulated training-free MoE-VLM expert merging around phase-conditioned expert roles, preserving the expert--phase information obscured by global routing aggregation.
We proposed \RoleMerge{}, a training-free method that represents phase-conditioned expert roles with Routing Role Profiles and uses expert--phase information loss to determine merge compatibility while preserving answer-decoding expert distinctions.
Across three models and six benchmarks, \RoleMerge{} retained more of the model's macro-average performance than the comparison methods at all evaluated expert-retention ratios, with relative gains of up to 9.6\%.
Together, these results validate phase-conditioned expert roles as a basis for training-free MoE-VLM expert merging under matched settings.

\noindent\textbf{Discussion.}
Routing Role Profiles are constructed from routing statistics on calibration data and characterize how experts are used across image-context processing, question processing, and answer decoding.
They should not be interpreted as universal semantic labels.
Our evaluation covers three models from two MoE-VLM families and three expert-retention ratios, leaving broader architectures, calibration domains, and open-ended generation settings for future study.
\RoleMerge{} focuses on merge compatibility and does not directly address parameter interference within merged experts.
Future work could combine \RoleMerge{} with more advanced parameter fusion or lightweight post-merge adaptation.
Beyond merging, expert pruning offers another way to reduce expert storage.
Extending phase-conditioned expert roles to pruning could help avoid removing experts whose routing is concentrated in inference phases with fewer tokens.

%% file: sections/appendix.tex
\section{Additional Technical Analysis}

\subsection{Non-Identifiability under Global Routing Aggregation}
The globally aggregated routing distribution
$\overline p_\ell(e)=\sum_b\widehat{\pi}_\ell(b)p_{\ell,b}(e)$
does not identify an expert's phase-conditioned role.
Consider the two-phase restriction with two equally likely phases and
two experts.
In a separated routing world, each phase exclusively uses a different expert; in a pooled world, both phases use both experts uniformly.
Both produce
$\overline p_\ell(e_1)=\overline p_\ell(e_2)=1/2$, whereas
\begin{equation}
I_{P_\ell}^{\mathrm{separated}}(S_\ell;B)=\log 2,
\qquad
I_{P_\ell}^{\mathrm{pooled}}(S_\ell;B)=0.
\end{equation}
Thus identical globally aggregated routing statistics are compatible
with either complementary or identical phase-conditioned expert roles.

\subsection{Derivation of the Expert--Phase Information Loss}
For a group $G=c_\ell^{-1}(g)$, conditioning on its compressed identity gives
\begin{equation}
P_\ell(B=b\mid C_\ell=g)
=q_{\ell,G}(b)
=\frac{\sum_{e\in G}P_\ell(e)q_{\ell,e}(b)}
{\sum_{e\in G}P_\ell(e)}.
\end{equation}
Because $C_\ell=c_\ell(S_\ell)$ is a deterministic function of
$S_\ell$,
\begin{align}
I_{P_\ell}(S_\ell;B)-I_{P_\ell}(C_\ell;B)
&=I_{P_\ell}(S_\ell;B\mid C_\ell)\nonumber\\
&=\sum_G\sum_{e\in G}P_\ell(e)
D_{\mathrm{KL}}(q_{\ell,e}\Vert q_{\ell,G}).
\end{align}
If only experts $i,j$ are merged, all other one-expert groups contribute zero.
Substituting $q_{\ell,\{i,j\}}=m_{\ell,ij}$ therefore yields exactly
the pairwise expert--phase information loss
$d_{\mathrm{EP},\ell}(i,j)$ defined in the main paper.
This establishes an exact pairwise expert--phase information loss, but does not imply that the partition produced by single linkage globally minimizes the group-level expert--phase information loss.
Nor does the identity claim that retaining more global phase mutual information universally predicts higher downstream accuracy.

\subsection{Derivation of Answer-Carrier Recoverability}
Fix an answer-decoding task--layer calibration record $r$ and grouping
$c_\ell$, let $p^A_r(e)$ be expert $e$'s normalized answer-decoding
routing mass, and let $G\in\mathcal G_\ell$ be a retained group.
After observing only group $G$, the Bayes-optimal guess is the member with largest $p^A_r(e)$.
Its conditional success probability is
\begin{equation}
q^A_r(G)
=\frac{\max_{e\in G}p^A_r(e)}
{\sum_{e\in G}p^A_r(e)}.
\end{equation}
Weighting by the probability of observing each group gives
\begin{equation}
\operatorname{ACR}_r(c_\ell)
=\sum_{G\in\mathcal G_\ell}\max_{e\in G}p^A_r(e).
\end{equation}
This is precisely
$P_{\mathrm{guess}}(S_\ell\mid C_\ell,B=\mathrm{answer},r)$.
Zero-mass groups contribute zero and are omitted from the conditional ratio.
The odds-form ambiguity $1/q^A_r(G)-1$ equals
$\sum_{e\in G}p^A_r(e)/\max_{e\in G}p^A_r(e)-1$, the unstabilized form of the group-normalized statistic underlying the signed collision adjustment.
For the audited record set $\mathcal R$, reported ACR is the uniform average $|\mathcal R|^{-1}\sum_{r\in\mathcal R}\operatorname{ACR}_r(c_r)$, where $c_r$ is the grouping at the layer associated with $r$.
ACR depends only on calibration routes and topology, not correctness labels.
We use it as a mechanism diagnostic rather than the greedy objective or a universal accuracy predictor.

\subsection{Additional Signed-Collision Diagnostic}

\begin{table}[H]
\centering
\small
\setlength{\tabcolsep}{2.5pt}
\begin{tabular}{@{}lrr>{\columncolor{avgcol}}r@{}}
\toprule
\textbf{Qwen3-VL, $\rho=.625$} & \textbf{ACR (\%)} & \textbf{EP ret.} & \textbf{Avg} \\
\midrule
\RoleMerge{} w/o collision adj. & 84.66 & \textbf{.8338} & 59.9 \\
\rowcolor{rrprow}
\RoleMerge{} ($\lambda=.03$) & \textbf{88.22} & .7752 & \textbf{69.2} \\
\bottomrule
\end{tabular}
\caption{Additional one-factor signed-collision diagnostic. ACR averages 144 task--layer calibration records from the task-disjoint TCS calibration set; EP ret.\ is the ratio of grouped to original expert--phase normalized mutual information summed over the same records, and Avg is the downstream unweighted six-task average.}
\label{tab:acr_collision_audit}
\end{table}

Enabling the signed collision adjustment raises displayed ACR by 3.56 percentage points and Avg
by 9.3 points even though the aggregated expert--phase information
retention decreases.
This additional intervention supports the adjustment's intended role
in preserving answer-decoding expert distinctions.
This is an expected trade-off because the full score balances role
compatibility against answer-decoding collision rather than optimizing
aggregated expert--phase information retention alone; it also reinforces
the main paper's treatment of ACR as a mechanism diagnostic rather than
a universal predictor of downstream performance.

\section{Implementation Details}

\subsection{Phase Boundaries and Routing Role Profile Construction}
\RoleMerge{} constructs its grouping evidence from routing traces of
the pretrained, frozen MoE-VLM.
For each calibration example, we run greedy autoregressive decoding with temperature set to 0 and sampling disabled, and partition routed assignments into three disjoint phase buckets.
The image-context bucket contains routed visual-token positions in the
prompt prefill and represents image-context processing; the question
bucket contains the remaining routed prompt positions and represents
question processing; and the answer bucket contains routed positions
produced during autoregressive decoding and represents answer decoding.
At layer $\ell$, each selected top-$k$ expert contributes one hard assignment to the corresponding bucket; gate probabilities are not used as fractional counts.
The resulting counts are $n_{\ell,b}(e)$ in the notation of the main
paper and are summed across calibration examples separately for each
inference phase.
Each phase column is then normalized over experts and assigned prior mass $1/3$, as defined in the main paper.
This yields the phase distributions $p_{\ell,b}(e)$, the balanced joint distribution $P_{\ell}(e,b)$, and the Routing Role Profile $q_{\ell,e}(b)$.
No task label, answer correctness, or downstream score is used when constructing these profiles.
All audited routing traces pass assignment-conservation and
nonempty-phase-bucket checks.
If an expert has zero balanced mass, its conditional RRP is represented by the zero vector and its contribution to each pairwise expert--phase information loss is zero.
If every pairwise expert--phase information loss in a layer is zero, the normalized pairwise-loss matrix remains all zero rather than dividing by zero.

\subsection{Answer-Aware Grouping and Parameter Merging}
For every compressed layer, the target number of experts is $K_\ell=\rho N_\ell$.
Pairwise expert--phase information losses $d_{\mathrm{EP},\ell}$ are normalized within each layer and extended to candidate groups through the group-wise role dissimilarity $d_{\mathrm{role},\ell}$, constructed using single linkage.
\RoleMerge{} then applies deterministic agglomeration using the signed collision adjustment in the main paper's merge score.
The default collision weight is $\lambda=.03$; the strict $\lambda=0$ control removes this adjustment.
The collision normalization scale $\tau_\ell$ is the median strictly
positive pairwise collision increment, with fallback value $1$ when
no such increment exists.
At each step, the implementation chooses the candidate with the
smallest merge score and resolves exact ties lexicographically by
group-wise role dissimilarity, signed collision adjustment,
merged-group size, and expert indices.
It repeats this update until exactly $K_\ell$ groups remain.

After the grouping map $c_\ell$ is fixed, the expert and router parameters in each group are merged using the same within-group weights as in the main paper.
For group $G$, the implementation pools each expert's nonnegative
calibration routing count over inference phases, then normalizes these
masses within $G$ into weights $\omega_{\ell,e}$ with
$\sum_{e\in G}\omega_{\ell,e}=1$.
It falls back to uniform weights only when the pooled mass of the entire group is zero.
The merged expert parameters $\widehat{\Theta}_{\ell,G}$ and
corresponding router entry $\widehat{\mathbf r}_{\ell,G}$ are both
computed with these same within-group weights.
Only routed expert tensors and router output entries are changed; the visual encoder, attention modules, shared parameters, and decoding interface remain unchanged.

\subsection{End-to-End Pseudocode}
Algorithm~\ref{alg:rolemerge_end_to_end} summarizes the complete
\RoleMerge{} pipeline.
Phase-separated hard routing counts determine the RRPs and the
answer-aware merge score, while the merged expert parameters and
corresponding router entry use the same normalized calibration-frequency
weights.

\input{sections/appendix_rolemerge_algorithm}

\section{Experimental Protocol}

\subsection{Models, Compared Methods, and Evaluation}
The main experiments cover DeepSeek-VL2-Tiny, DeepSeek-VL2-Small~\cite{deepseekvl22024}, and Qwen3-VL-30B-A3B-Instruct~\cite{qwen3vl2025}.
The Full Model is the uncompressed reference.
Sub-MoE~\cite{submoe2025} and NAMEx~\cite{namex2026} are evaluated
across all three models; MC-SMoE~\cite{mcsmoe2024},
MergeMoE~\cite{mergemoe2025}, HC-SMoE~\cite{hcsmoe2025}, and
REAM~\cite{ream2026} are additionally evaluated on
DeepSeek-VL2-Tiny.
At the same model and expert-retention ratio, all compressed methods
use the same checkpoint, calibration set, retained expert count, and
compressed MoE layers.
Unless otherwise stated, experiments use the TCS calibration set
described below.

Evaluation follows the main paper.
InfoVQA~\cite{infovqa2022}, MMMU~\cite{mmmu2024}, and OCRBench~\cite{ocrbench2023} form MM3; ARC-Easy~\cite{arc2018}, PIQA~\cite{piqa2020}, and Winogrande~\cite{winogrande2020} form Text3.
All task values are official task scores mapped to a 0--100 scale: InfoVQA uses ANLS$\times100$, while the other five tasks use their official accuracy$\times100$.
MM3 and Text3 are the unweighted averages over their respective three
tasks, and Avg is the unweighted average over all six.
Unless otherwise stated, experiments use two NVIDIA A100 80GB GPUs;
the isolated efficiency audit below uses one A100 80GB.

\subsection{Evaluation Identities, Prompts, and Decoding}
Table~\ref{tab:evaluation_protocol} gives the exact paper-facing task identifiers and evaluated sample counts.
The evaluation cap is 1,000 examples per task; InfoVQA and MMMU contain only 500 and 900 examples, respectively, under the archived validation-task definitions, so all available examples are used.
For every model, all methods use the same fixed example identities and order within a task.
The multimodal tasks use pinned LMMS-Eval task definitions and the text tasks use pinned LM Evaluation Harness definitions.
Their task templates, answer extraction, metric implementation, checkpoint-specific processor, and generation arguments are held fixed across methods; only the compressed checkpoint changes.
All autoregressive generation uses greedy decoding with temperature set to 0 and sampling disabled.
We use batch size one for the audited generation runs.
For Qwen3-VL, the archived runner uses bfloat16, tensor parallelism of two, a 32,768-token maximum context, and the same processor bounds and system prompt for every compared method.
The DeepSeek-VL2-Small comparison likewise shares one dtype-safe wrapper, processor, evaluator, prompt set, and generation configuration across all rows.

\begin{table}[!htbp]
\centering
\small
\setlength{\tabcolsep}{1.5pt}
\begin{tabular}{@{}llrl@{}}
\toprule
\textbf{Benchmark} & \textbf{Archived task identifier} & \textbf{$n$} & \textbf{Paper score} \\
\midrule
InfoVQA & \texttt{infovqa\_val\_lite} & 500 & ANLS$\times100$ \\
MMMU & \texttt{mmmu\_val} & 900 & Accuracy$\times100$ \\
OCRBench & \texttt{ocrbench} & 1,000 & Accuracy$\times100$ \\
ARC-Easy & \texttt{arc\_easy} & 1,000 & Accuracy$\times100$ \\
PIQA & \texttt{piqa} & 1,000 & Accuracy$\times100$ \\
Winogrande & \texttt{winogrande} & 1,000 & Accuracy$\times100$ \\
\bottomrule
\end{tabular}
\caption{Evaluation identities and sample counts used by every paper-facing comparison. The same identities are reused across methods within each model and retention ratio.}
\label{tab:evaluation_protocol}
\end{table}

\subsection{Matched-Baseline Construction}
The reported baseline numbers are reruns under our common MoE-VLM protocol, not values copied from the original papers.
At a fixed model and $\rho$, every method receives the same base checkpoint, target expert counts $K_\ell$, compressed layer set, calibration identities, evaluation identities, and hardware precision.
Each method retains its defining grouping and parameter-combination
rules. The shared adaptation is limited to loading the same MoE-VLM
checkpoint, materializing the requested number of retained experts,
translating the router to the reduced expert pool, and exposing the
checkpoint through the same evaluator.
Thus ``matched'' does not mean replacing a baseline's grouping or
parameter-merging procedure with that of \RoleMerge{}.

For DeepSeek-VL2-Small, one TCS calibration set with 64 examples from
each source is reused by \RoleMerge{}, Sub-MoE, and NAMEx at all three
retention ratios.
The three rows share task configurations, prompts, sample identities, processor, evaluator, and generation settings.
For DeepSeek-VL2-Tiny, the main TCS calibration set uses 128 examples
per source.
The Qwen3-VL audit checks protocol parity per task and retention ratio, including evaluator version, precision, tensor parallelism, context length, batch size, prompts, processor bounds, and sample count.
No method is selected separately per downstream task, and the uncompressed Full Model is evaluated once per model and reused as the common reference across retention ratios.
Because some original baselines were proposed for text-only MoEs, these rows should be interpreted as matched MoE-VLM adaptations under the paper's protocol rather than reproductions of the original papers' task settings.

\subsection{TCS Calibration Set}
TCS denotes TextVQA~\cite{singh2019textvqa}, ChartQA~\cite{masry2022chartqa}, and ScienceQA~\cite{lu2022scienceqa}.
These sources provide scene-text reading, chart/number understanding, and scientific visual reasoning contexts for route profiling.
They are used only to collect calibration routing traces, not as
downstream supervision.
\RoleMerge{} uses no task labels, answer correctness, or downstream
evaluation scores when constructing RRPs from this set.
All audited routing traces pass assignment-conservation and nonempty
phase-bucket checks.

\section{Additional Results for Q1}

\subsection{REAM Port and Additional REAP Baseline}
Table~\ref{tab:reap_ream_tiny} reports detailed REAM and REAP results on DeepSeek-VL2-Tiny.
REAM~\cite{ream2026} is included in the main comparison, while the table here retains two-decimal scores and juxtaposes it with REAP.
REAP~\cite{reap2025} removes low-saliency experts rather than merging expert parameters, so its compression operator differs from the expert-merging methods in the main comparison.
Because the official REAM implementation does not natively support
DeepSeek-VL2, its results use our audited official-core DeepSeek-VL2
port with non-sequential merging; the calibration replay does not
retain the complete visual sequence and attention state required by
REAM's sequential mode.
The REAP adapter follows its router-weighted expert-activation criterion, and all calibration identities remain disjoint from the six evaluation datasets.

\paragraph{Retention trend and operator comparison.}
Both supplementary operators improve smoothly as more experts are
retained.  From $\rho=.50$ to $.75$, six-task Avg rises by 4.94 points
for REAM and 4.54 points for REAP; the corresponding MM3 gains are
5.16 and 4.51 points, and the Text3 gains are 4.73 and 4.57 points.
This monotone response is useful as a basic implementation check:
neither port exhibits an anomalous reversal when its retained expert
budget increases.

\input{sections/appendix_reap_ream_table}

The two operators remain close under the matched protocol.
REAP exceeds REAM by 0.27 and 0.80 Avg points at $\rho=.50$ and $.625$,
whereas REAM is higher by 0.13 points at $\rho=.75$.
Thus the supplementary pruning baseline does not uniformly dominate
the merging baseline, and their largest displayed separation is below
one point.
Both remain below the uncompressed Full Model, including gaps of 4.92
and 5.05 Avg points for REAM and REAP at $\rho=.75$, respectively.
These comparisons describe the compression trade-off without treating
the two distinct operators as interchangeable.

\subsection{Paired Uncertainty}

Table~\ref{tab:small_tcs_bootstrap} complements the main Q1 results with paired uncertainty for the matched DeepSeek-VL2-Small comparisons.
These intervals quantify finite-sample evaluation uncertainty rather than variation from repeated decoding runs.
For each task, one bootstrap replicate samples evaluation identities with replacement and applies exactly the same sampled indices to both methods.
We compute each task mean, average the six task means with equal weight, and record the paired difference.
The reported interval is the 2.5th--97.5th percentile range over 10,000 replicates.
This preserves pairing between methods while preventing evaluation
sets with more examples from receiving greater weight in the six-task
Avg.

\par\smallskip
\noindent\begin{minipage}{\columnwidth}
\centering
\small
\setlength{\tabcolsep}{2pt}
\begin{tabular}{@{}llrr@{}}
\toprule
\textbf{$\rho$} & \textbf{Comparison} & \textbf{$\Delta$} & \textbf{95\% CI} \\
\midrule
.50 & \RoleMerge{} $-$ Sub-MoE & $+2.84$ & $[+1.54,+4.15]$ \\
.50 & \RoleMerge{} $-$ NAMEx & $+3.10$ & $[+1.82,+4.42]$ \\
.625 & \RoleMerge{} $-$ Sub-MoE & $+3.53$ & $[+2.32,+4.72]$ \\
.625 & \RoleMerge{} $-$ NAMEx & $+3.69$ & $[+2.51,+4.88]$ \\
.75 & \RoleMerge{} $-$ Sub-MoE & $+2.21$ & $[+1.13,+3.30]$ \\
.75 & \RoleMerge{} $-$ NAMEx & $+2.48$ & $[+1.40,+3.59]$ \\
\bottomrule
\end{tabular}
\captionof{table}{Paired 10,000-resample DeepSeek-VL2-Small differences in six-task Avg points. Resampling uses matched per-sample scores.}
\label{tab:small_tcs_bootstrap}
\end{minipage}
\par\medskip

Table~\ref{tab:qwen_bootstrap} gives the corresponding paired
uncertainty for the Qwen3-VL-30B-A3B-Instruct comparison with Sub-MoE.

\input{results/aaai27_qwen_closure_20260713/qwen3vl_supplement_bootstrap}

\section{Additional Mechanism Analysis for Q2}

\subsection{Component Controls}
Table~\ref{tab:small_grouping_ablation} reports the DeepSeek-VL2-Small component controls summarized in the main paper at $\rho=.625$.
All rows use the same calibration set, retained expert count, grouping
rule, and expert/router parameter-merging procedure; only the specified
mechanism changes.
Pooling routing statistics from image-context processing, question
processing, and answer decoding before grouping reduces ACR from
95.0\% to 93.1\% and MM3 from 57.1 to 53.5.
Removing the signed collision adjustment lowers ACR to 89.1\% and MM3
to 55.4, while permuting RRPs across expert identities lowers them to
70.8\% and 51.7, respectively.
The corresponding paired 95\% intervals exclude zero.
The table uses the complete \RoleMerge{} row from the main text as its
reporting anchor: paired ablation deltas are computed before display
rounding and applied to that common anchor.

\input{sections/appendix_small_ablation_table}

\section{Additional Calibration Analyses for Q3}

\subsection{Calibration Construction, Source Type, and Set Size}
Table~\ref{tab:calibration_controls} expands the calibration-set
construction and source-type controls summarized under Q3 in the main
paper.
For the source-type comparison, the model, 128 examples per source,
compression settings, three retention ratios, and six evaluation tasks
are fixed; only the calibration source used to construct the RRPs
changes.
The mixed-multimodal set contains image--question--answer traces, the
text-only-answer set contains answer traces without visual conditioning,
and the task-specific-multimodal set uses one task-specific multimodal
source rather than the mixed pool.
The first two rows compare TCS with the earlier Mixed/IQA calibration
construction.
The results show that mixed multimodal calibration is stronger than the
two source-type controls, while TCS is close to the alternative
multimodal construction over the three-retention-ratio sweep.

\input{sections/appendix_calibration_controls_table}

\paragraph{Construction and source-type contrasts.}
TCS and Mixed/IQA are nearly tied on the three-ratio mean:
TCS is higher by 0.09 points overall, with per-ratio differences of
$+0.29$, $-0.62$, and $+0.59$.
The change in sign is consistent with treating TCS as a fixed,
balanced construction rather than a calibration set optimized for a
particular retention ratio.
The source-type control is more decisive.
Mixed multimodal calibration exceeds text-only-answer calibration by
6.48 mean points and the task-specific multimodal control by 4.57
points.
These gaps are already present at every retention ratio, supporting
the use of phase-complete multimodal traces while avoiding a claim
that any individual source mixture is downstream-optimal.

\begin{table}[H]
\centering
\small
\setlength{\tabcolsep}{1.2pt}
\begin{tabular}{@{}lrrrrrr@{}}
\toprule
\textbf{Examples/source} & \textbf{32} & \textbf{64} & \textbf{128} & \textbf{256} & \textbf{512} & \textbf{1024} \\
\midrule
Mean over $\rho$ & 54.88 & 56.79 & 57.63 & 56.70 & 57.41 & 57.18 \\
\bottomrule
\end{tabular}
\caption{DeepSeek-VL2-Tiny calibration-set-size sweep (six-task Avg points).}
\label{tab:profile_size}
\end{table}

Table~\ref{tab:profile_size} varies the number of calibration examples
per source.
The largest change is the 2.75-point increase from 32 to 128 examples
per source.
From 64 through 1024 examples, however, all six-task averages lie
within a 0.93-point range; increasing the set from 128 to 1024 changes
the mean by only $-0.45$ points.
The non-monotone response therefore forms a broad plateau rather than
a sample-count scaling trend.
It also shows that the reported behavior is not explained by selecting
one sharply tuned calibration-set size.

\input{Figures/fig_calibration_diagnostics}

\subsection{Fixed-Budget Mixtures and Routing Coverage}
To test sensitivity to the composition of TCS, we additionally evaluate
fixed-budget calibration sets on DeepSeek-VL2-Tiny at $\rho=.625$.
The model, \RoleMerge{} configuration, phase buckets, total calibration
budget, and InfoVQA/MMMU evaluation identities are fixed; only source
count, source prior, or source identity changes.
TCS was selected before inspecting these downstream diagnostic scores.

The exact-mixture comparison separates downstream score from the
calibration-only routing criterion: DAR reaches 42.05 on this two-task
diagnostic, compared with 39.69 for TCS.
Thus neither the main paper nor this appendix claims downstream
optimality for TCS.

\input{sections/appendix_calibration_mixture_tables}

\paragraph{Budget-matched source-count comparison.}
Because every row uses 384 calibration examples, the first panel
isolates composition rather than total calibration volume.
The three one-source sets span only 0.50 points (38.86--39.36) and
average 39.04.
The three two-source sets have a similar mean of 39.76 but a wider
1.98-point range: T+C and T+S reach 40.33 and 40.47, whereas C+S reaches
38.49.
Within this audit, adding a second source therefore does not provide a
uniform gain; source identity matters at least as much as source count.
The balanced three-source TCS value of 39.69 is 0.65 points above the
one-source mean, 0.07 points below the two-source mean, and 0.78 points
below the strongest individual two-source row.
This places TCS near the center of the fixed-budget source-count
comparisons rather than at an isolated score extreme.

\paragraph{Exact-mixture spread.}
Holding the same total budget and source count while changing the exact
source identities produces a larger score spread.
DCS and TDC exceed TCS by 0.61 and 0.87 points, while DAR exceeds it by
2.36 points.
DAR is also 1.75 points above DCS and 1.49 points above TDC.
These differences show why the exact source identities must be reported
alongside the nominal number of sources.
They do not alter the matched main comparisons, which reuse the same
pre-specified TCS set for every method and retention ratio.

\paragraph{Within-panel identity pattern.}
Among the one-source controls, T reaches 39.36, which is 0.48 points
above the mean of C and S.
Among the two-source controls, the two mixtures containing T average
40.40, or 1.91 points above C+S.
This repeated ordering is descriptive rather than causal, because
holding the total budget fixed also changes the number of examples
allocated to each included source.
It nevertheless explains why the mean over source counts can conceal
material composition effects.
The 2.36-point range among the exact three-source mixtures also slightly
exceeds the 1.98-point range in the entire source-count panel, further
motivating the identity-level audit.

\paragraph{Source-count and routing-coverage diagnostics.}
Figure~\ref{fig:calibration_diagnostics}(c) shows that one-source, two-source, and TCS
calibration sets give close two-task averages of 39.04, 39.76, and
39.69.
Instead, Table~\ref{tab:calibration_route_coverage} shows why TCS is
kept as a pre-specified, balanced default: among the audited mixtures,
it has the lowest top-16 answer-decoding expert overlap and the highest
union coverage.
The score diagnostic bounds the optimality claim, while the routing
diagnostic supports the fixed default used by every matched main
comparison.
More specifically, relative to TCS, DCS increases $J_{16}$ by .042 and
reduces $U_{16}$ by .017; the corresponding changes are $+.035$ and
$-.009$ for DAR, and $+.096$ and $-.033$ for TDC.
Thus all three alternative exact mixtures show greater top-16 overlap
and lower union coverage than TCS.
Across these four rows, the ordering of $J_{16}$ is exactly reversed by
the ordering of $U_{16}$, making the two columns complementary views of
the same concentration--coverage trade-off.

\input{sections/appendix_calibration_route_coverage_table}

\paragraph{Score--coverage separation.}
The downstream score ordering does not follow either routing statistic
monotonically.
DAR has the highest two-task Avg but only the second-highest union
coverage, while TCS has the highest union coverage without the highest
Avg.
TDC provides the clearest counterexample to treating overlap as an
accuracy proxy: it has the largest $J_{16}$ and smallest $U_{16}$, yet
its Avg remains between DCS and DAR.
We therefore use these columns only to audit the intended routing
diversity of the fixed calibration default, not to predict downstream
accuracy.
The diagnostic is limited to one checkpoint, one retention ratio, and
the InfoVQA/MMMU pair; it supports protocol consistency but does not
claim that TCS is downstream-optimal for other models or task sets.

\subsection{Collision Weight, Grouping Rule, and Search Depth}
\input{sections/appendix_sensitivity_table}
\input{results/aaai27_solver_only_20260717/audit/solver_only_table}

Table~\ref{tab:sensitivity} provides the two-decimal values underlying
the Q3 sensitivity summary in the main paper.
Increasing $\lambda$ from .03 to .10 changes Avg by $+0.64$, while removing the signed collision adjustment drops Avg by 2.55.
The small gain at .10 rules out an optimality claim for .03, while the
larger drop at zero supports a positive weight over this audited range.
The grouping-rule block shows a clearer conclusion: single linkage
outperforms average linkage, complete linkage, and five-seed $k$-means++ when
the same expert--phase information loss and parameter-merging procedure
are used.

\paragraph{Metric-wise sensitivity.}
The collision-weight sweep is not strictly monotone, which is why we
treat $\lambda=.03$ as a fixed default rather than a tuned optimum.
Relative to this default, $\lambda=.01$ raises MM3 by 0.51 points but
lowers Text3 by 1.23 points, producing the modest 0.36-point Avg
decrease.
At $\lambda=.10$, the corresponding changes are $+0.58$ and $+0.70$,
whereas setting $\lambda=0$ reduces both metrics by 3.23 and 1.86
points.
Thus the audit supports retaining a positive collision term, while the
small differences among the positive settings do not justify a claim
that one precise weight is universally best.

\paragraph{Variation across positive weights.}
Restricting the comparison to $\lambda\in\{.01,.03,.10\}$ gives ranges
of 0.58 points on MM3, 1.93 points on Text3, and 1.00 point on Avg.
Including $\lambda=0$ expands the Avg range to 3.19 points.
The main $\lambda=.03$ setting lies inside the positive-weight band:
its Avg is 0.36 points above .01 and 0.64 points below .10.
Accordingly, the larger distinction in this sweep is between removing
the adjustment and retaining a positive adjustment, not between two
nearby positive values.
This is a local stability observation, not a hyperparameter-optimality
claim.

\paragraph{Grouping-rule decomposition.}
Average and complete linkage trail single linkage on both displayed
metrics.
Their losses are larger on Text3 (1.70 and 2.20 points) than on MM3
(0.41 and 1.08 points), so their lower Avg is not produced by a reversal
between the two diagnostic aggregates.
The $k$-means++ control has a different profile: its 5.44-point Avg gap
is driven mainly by a 9.65-point MM3 deficit, while its Text3 deficit is
1.33 points.
Even the displayed mean plus one standard deviation for its Avg remains
4.24 points below single linkage.
These results favor the deterministic default based on single linkage for this
matched diagnostic without implying that it must dominate every
possible clustering rule or evaluation setting.

Table~\ref{tab:solver_only} gives the solver-depth diagnostic.
Panel A changes search depth under a fixed answer-aware merge score,
while Panel B compares grouping rules at $\rho=.625$.
Lower merge-score values do not automatically translate into better
downstream performance, consistent with the main paper's use of the
score as a greedy grouping criterion rather than a universal accuracy
predictor.
The local-search variants improve Avg only at $\rho=.625$ and decline
at both $.50$ and $.75$, so they are not a consistently stronger
replacement for the fixed greedy procedure.

\section{Storage and Efficiency Details for Q4}

\subsection{DeepSeek-VL2-Small Single-GPU Efficiency}
\input{sections/appendix_small_efficiency_table}

Table~\ref{tab:small_single_gpu_efficiency} reports an isolated single-A100 measurement for the Full Model and all three \RoleMerge{} retention ratios.
All four checkpoints remain fully GPU-resident.
At $\rho=.50$, \RoleMerge{} reduces parameters by 44.6\%, steady-state incremental memory by 33.8\%, and peak incremental memory by 32.2\% relative to Full.
Its median time to first token decreases by 32.1\%, normal end-to-end latency decreases by 15.3\%, and fixed-length decoding throughput increases by 5.7\%.
At $\rho=.625$, the corresponding parameter, steady-memory, and peak-memory reductions are 33.4\%, 25.4\%, and 24.1\%.
The $\rho=.50$ latency and throughput directions hold in all three repetitions.
For $\rho=.625$, one of three normal-generation repetitions does not improve end-to-end latency, so we report its aggregate timing without claiming a stable speedup.
The modest decoding-throughput differences should be interpreted as system measurements rather than algorithmic asymptotics.

\subsection{Exact Footprint Accounting}
\begin{center}
\begin{minipage}{\columnwidth}
\centering
\small
\setlength{\tabcolsep}{1.2pt}
\begin{tabular}{@{}lrrr@{}}
\toprule
\shortstack[l]{\textbf{Qwen3-VL-30B-A3B-}\\\textbf{Instruct artifact}} & \textbf{Full} & \textbf{$\rho=.625$} & \textbf{Retained} \\
\midrule
Routed-expert parameters & 28.991B & 18.119B & 62.5\% \\
Total parameters & 31.071B & 20.194B & 65.0\% \\
Serialized bf16 weights & 62.14GB & 40.39GB & 65.0\% \\
\bottomrule
\end{tabular}
\captionof{table}{Qwen3-VL-30B-A3B-Instruct parameter and checkpoint accounting from tensor headers.}
\label{tab:qwen_footprint}
\end{minipage}
\end{center}

The retention ratio $\rho$ applies to routed experts.
At $\rho=.625$, Qwen3-VL-30B-A3B-Instruct routed-expert parameters decrease from 28.991B to 18.119B, while total parameters and serialized bf16 weights both retain 65.0\% of the full model.

%% file: sections/appendix_rolemerge_algorithm.tex
\begin{algorithm*}[p]
\caption{\RoleMerge{}: Phase-Conditioned Grouping and Expert--Router Parameter Merging}
\label{alg:rolemerge_end_to_end}
\footnotesize
\begin{algorithmic}[1]
\Require Pretrained MoE-VLM $F$; calibration set
$\mathcal D_{\mathrm{cal}}$ (labels unused); selected MoE layers
$\mathcal L$; expert-retention ratio $\rho$; collision weight
$\lambda=.03$
\Ensure Compressed model $\widehat F$ with $K_\ell$ routed experts in each compressed layer $\ell$

\Statex \textbf{Phase I: collect phase-separated routing evidence}
\ForAll{$x\in\mathcal D_{\mathrm{cal}}$}
    \State Run deterministic autoregressive inference with the frozen model $F$
    \State Split routed positions into image-context processing,
    question processing, and answer decoding
    \State For each $\ell\in\mathcal L$, phase $b$, and top-$k$
    selected expert $e$, increment $n_{\ell,b}(e)$
\EndFor

\Statex \textbf{Phase II: construct RRPs and pairwise expert--phase information losses}
\ForAll{$\ell\in\mathcal L$}
    \ForAll{phases $b$}
        \State $p_{\ell,b}(e)\gets
        n_{\ell,b}(e)/\sum_j n_{\ell,b}(j)$
    \EndFor
    \State $P_\ell(e,b)\gets p_{\ell,b}(e)/3$;
    $P_\ell(e)\gets\sum_bP_\ell(e,b)$;
    $q_{\ell,e}(b)\gets P_\ell(e,b)/P_\ell(e)$, using $q_{\ell,e}\equiv0$ when $P_\ell(e)=0$
    \ForAll{expert pairs $i<j$}
        \If{$P_\ell(i)+P_\ell(j)>0$}
        \State $m_{\ell,ij}\gets
        \bigl(P_\ell(i)q_{\ell,i}+P_\ell(j)q_{\ell,j}\bigr)/
        \bigl(P_\ell(i)+P_\ell(j)\bigr)$
        \State $d_{\mathrm{EP},\ell}(i,j)\gets
        P_\ell(i)D_{\mathrm{KL}}(q_{\ell,i}\Vert m_{\ell,ij})
        +P_\ell(j)D_{\mathrm{KL}}(q_{\ell,j}\Vert m_{\ell,ij})$
        \Else
        \State $d_{\mathrm{EP},\ell}(i,j)\gets0$
        \EndIf
    \EndFor
    \State Normalize $d_{\mathrm{EP},\ell}$ by its largest off-diagonal value,
    retaining an all-zero matrix when the maximum is zero
    \State $\kappa_\ell\gets1-H(p_{\ell,\mathrm{ans}})/\log N_\ell$
\EndFor
\State $\bar\kappa\gets|\mathcal L|^{-1}\sum_{\ell\in\mathcal L}
\max(\kappa_\ell,\varepsilon)$

\Statex \textbf{Phase III: construct the answer-aware grouping}
\ForAll{$\ell\in\mathcal L$}
    \State $\lambda_\ell\gets\lambda\max(\kappa_\ell,\varepsilon)/\bar\kappa$
    \State $K_\ell\gets\Call{TargetCount}{N_\ell,\rho}$;
    $\mathcal G_\ell\gets\{\{1\},\ldots,\{N_\ell\}\}$
    \State $R_{\mathrm{ans},\ell}(G)\gets
    \left[\sum_{e\in G}p_{\ell,\mathrm{ans}}(e)/
    \bigl(\max_{e\in G}p_{\ell,\mathrm{ans}}(e)+\varepsilon\bigr)-1\right]_+$
    \State $\tau_\ell\gets\operatorname{median}^{+}_{i<j}
    \bigl(R_{\mathrm{ans},\ell}(\{i,j\})
    -R_{\mathrm{ans},\ell}(\{i\})
    -R_{\mathrm{ans},\ell}(\{j\})\bigr)$,
    using $1$ if no positive value exists
    \While{$|\mathcal G_\ell|>K_\ell$}
        \ForAll{unordered group pairs $G,H\in\mathcal G_\ell$}
            \State $d_{\mathrm{role},\ell}(G,H)\gets
            \min_{i\in G,j\in H}d_{\mathrm{EP},\ell}(i,j)$
            \State $\Delta R_{\mathrm{ans},\ell}(G,H)\gets
            R_{\mathrm{ans},\ell}(G\cup H)
            -R_{\mathrm{ans},\ell}(G)-R_{\mathrm{ans},\ell}(H)$
            \State $J_\ell(G,H)\gets d_{\mathrm{role},\ell}(G,H)
            +\lambda_\ell\Delta R_{\mathrm{ans},\ell}(G,H)/\tau_\ell$
        \EndFor
        \State Select $(G^\star,H^\star)\gets\arg\min J_\ell(G,H)$,
        breaking exact ties by group-wise role dissimilarity, signed collision adjustment, group size, and expert indices
        \State Replace $G^\star,H^\star$ in $\mathcal G_\ell$ by $G^\star\cup H^\star$
    \EndWhile

\Statex \textbf{Phase IV: merge expert and router parameters}
    \ForAll{$G\in\mathcal G_\ell$}
        \State Let $u_{\ell,e}\gets\sum_b n_{\ell,b}(e)$ and
        $\omega_{\ell,e}\gets
        u_{\ell,e}/\sum_{j\in G}u_{\ell,j}$ for $e\in G$
        \State If $\sum_{j\in G}u_{\ell,j}=0$, set
        $\omega_{\ell,e}\gets1/|G|$
        \State $\widehat{\Theta}_{\ell,G}\gets
        \sum_{e\in G}\omega_{\ell,e}\Theta_{\ell,e}$;
        $\widehat{\mathbf r}_{\ell,G}\gets
        \sum_{e\in G}\omega_{\ell,e}\mathbf r_{\ell,e}$
    \EndFor
    \State Replace layer $\ell$ by
    $\{(\widehat{\Theta}_{\ell,G},
    \widehat{\mathbf r}_{\ell,G}):G\in\mathcal G_\ell\}$ and reduce
    its router output dimension to $K_\ell$
\EndFor
\State \Return $\widehat F$
\end{algorithmic}
\end{algorithm*}

%% file: sections/appendix_reap_ream_table.tex
\begin{table*}[!t]
\centering
\small
\setlength{\tabcolsep}{3.2pt}
\begin{tabular}{@{}lrrrrrrrr>{\columncolor{avgcol}}r@{}}
\toprule
\textbf{Method / $\rho$} &
\textbf{Info} & \textbf{MMMU} & \textbf{OCR} & \textbf{MM3} &
\textbf{ARC-E} & \textbf{PIQA} & \textbf{Wino.} & \textbf{Text3} &
\textbf{Avg} \\
\midrule
\rowcolor{modelrow}
Full & 49.28 & 39.11 & 81.10 & 56.50 & 67.90 & 73.20 & 59.90 & 67.00 & 61.75 \\
\midrule
REAM / .50  & 40.31 & 34.11 & 70.20 & 48.21 & 46.40 & 67.30 & 53.00 & 55.57 & 51.89 \\
REAM / .625 & 43.95 & 34.78 & 72.90 & 50.54 & 51.80 & 68.80 & 57.40 & 59.33 & 54.94 \\
REAM / .75  & 46.46 & 35.44 & 78.20 & 53.37 & 54.00 & 69.00 & 57.90 & 60.30 & 56.83 \\
\midrule
REAP / .50  & 41.18 & 34.00 & 71.30 & 48.83 & 46.90 & 66.60 & 53.00 & 55.50 & 52.16 \\
REAP / .625 & 45.72 & 34.44 & 73.90 & 51.35 & 53.20 & 69.80 & 57.40 & 60.13 & 55.74 \\
REAP / .75  & 46.07 & 35.44 & 78.50 & 53.34 & 53.60 & 69.10 & 57.50 & 60.07 & 56.70 \\
\bottomrule
\end{tabular}
\caption{Detailed REAM~\cite{ream2026} merging and additional REAP~\cite{reap2025} pruning results on DeepSeek-VL2-Tiny using official task scores on a 0--100 scale. Both use the same TCS calibration set, three expert-retention ratios, and six-task evaluation protocol as the Tiny comparison in the main paper. REAM also appears in the main table and uses our audited official-core DeepSeek-VL2 port with non-sequential merging; REAP is supplementary because it removes experts rather than merging expert parameters. Avg is the unweighted six-task average.}
\label{tab:reap_ream_tiny}
\end{table*}

%% file: results/aaai27_qwen_closure_20260713/qwen3vl_supplement_bootstrap.tex
\par\smallskip
\noindent\begin{minipage}{\columnwidth}
\centering
\small
\begin{tabular}{llrr}
\toprule
\textbf{$\rho$} & \textbf{Comparator} & \textbf{$\Delta$ Avg} & \textbf{Paired 95\% CI} \\
\midrule
0.50 & Sub-MoE & $+5.97$ & $[+4.58,+7.32]$ \\
0.625 & Sub-MoE & $+4.68$ & $[+3.42,+5.90]$ \\
0.75 & Sub-MoE & $+2.82$ & $[+1.71,+3.97]$ \\
\bottomrule
\end{tabular}
\captionof{table}{Paired 10,000-resample uncertainty in six-task Avg points for the Qwen3-VL-30B-A3B-Instruct \RoleMerge{}--Sub-MoE comparisons reported in the main table.}
\label{tab:qwen_bootstrap}
\end{minipage}
\par\smallskip

%% file: sections/appendix_small_ablation_table.tex
\begin{table}[H]
\centering
\footnotesize
\setlength{\tabcolsep}{2pt}
\begin{tabular}{@{}lrrr@{}}
\toprule
\textbf{Intervention} & \textbf{ACR (\%)} & \textbf{MM3} & \textbf{95\% CI} \\
\midrule
w/o phase split & 93.1 & 53.5 & $[-5.46,-1.81]$ \\
w/o collision adj. & 89.1 & 55.4 & $[-3.27,-0.14]$ \\
permuted roles & 70.8 & 51.7 & $[-7.40,-3.62]$ \\
\rowcolor{rrprow}
Complete \RoleMerge{} & \textbf{95.0} & \textbf{57.1} & --- \\
\bottomrule
\end{tabular}
\caption{DeepSeek-VL2-Small component controls at $\rho=.625$. ACR is averaged uniformly over 78 task--layer calibration records; MM3 is the unweighted InfoVQA/MMMU/OCRBench average. The final column gives paired 10,000-resample 95\% intervals for the MM3 change relative to the complete \RoleMerge{} setting.}
\label{tab:small_grouping_ablation}
\end{table}

%% file: sections/appendix_calibration_controls_table.tex
\begin{table}[H]
\centering
\small
\setlength{\tabcolsep}{3.2pt}
\begin{tabular}{@{}lrrrr@{}}
\toprule
\textbf{Calibration construction} & \textbf{.50} & \textbf{.625} & \textbf{.75} & \textbf{Mean} \\
\midrule
Mixed/IQA, 64/source & 52.99 & 57.56 & 59.82 & 56.79 \\
\rowcolor{rrprow}
TCS, 128/source & 53.28 & 56.94 & 60.41 & 56.88 \\
\midrule
Mixed multimodal, 128/source & 54.05 & 58.19 & 60.65 & 57.63 \\
Text-only answer, 128/source & 45.08 & 51.41 & 56.95 & 51.15 \\
Task-specific MM, 128/source & 48.85 & 53.52 & 56.81 & 53.06 \\
\bottomrule
\end{tabular}
\caption{DeepSeek-VL2-Tiny calibration controls in six-task Avg points. The first two rows compare calibration-set constructions; the last three compare calibration-source types.}
\label{tab:calibration_controls}
\end{table}

%% file: Figures/fig_calibration_diagnostics.tex
\begin{figure*}[t]
  \centering
  \begin{minipage}[b]{0.318\textwidth}
    \centering
    \includegraphics[
      width=\linewidth,
      trim=0 0 216 0,
      clip
    ]{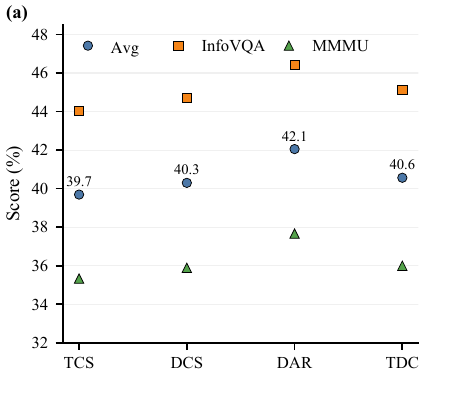}
  \end{minipage}\hfill
  \begin{minipage}[b]{0.318\textwidth}
    \centering
    \includegraphics[
      width=\linewidth,
      trim=216 0 0 0,
      clip
    ]{Figures/fig_calibration_mixture_diagnostic.pdf}
  \end{minipage}\hfill
  \begin{minipage}[b]{0.345\textwidth}
    \centering
    \includegraphics[
      width=\linewidth
    ]{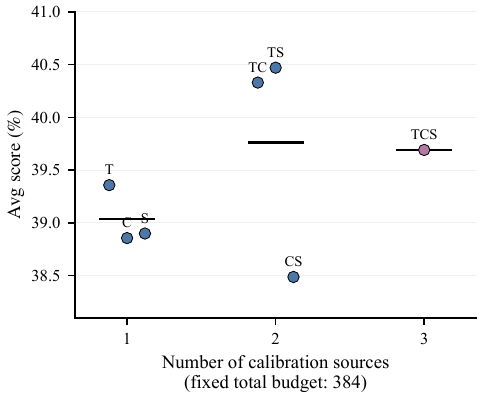}%
    \makebox[0pt][l]{%
      \hspace{-\linewidth}%
      \raisebox{0.75\linewidth}[0pt][0pt]{%
        \hspace{0.01\linewidth}{\small\bfseries(c)}%
      }%
    }%
  \end{minipage}
  \caption{Fixed-budget calibration diagnostics on DeepSeek-VL2-Tiny at
  $\rho=.625$. T/C/S/D/A/R denote TextVQA, ChartQA, ScienceQA, DocVQA,
  AI2D, and RealWorldQA. Avg is the unweighted mean of InfoVQA and MMMU.
  (a) Exact-mixture downstream scores. (b) Downstream Avg versus top-16
  answer-decoding union coverage; DAR has the highest audited score,
  whereas TCS has the highest coverage. (c) Source-count comparison under
  a fixed 384-example budget; points enumerate all one- and two-source
  subsets and the three-source TCS set, and horizontal bars mark
  within-count means.}
  \label{fig:calibration_diagnostics}
\end{figure*}

%% file: sections/appendix_calibration_mixture_tables.tex
\par\smallskip
\noindent\begin{minipage}{\columnwidth}
\centering
\small
\setlength{\tabcolsep}{3pt}
\begin{tabular}{@{}llrr@{}}
\toprule
\textbf{Panel} & \textbf{Calibration set} & \textbf{Budget} & \textbf{Avg} \\
\midrule
\multirow{7}{*}{Source count}
& T & 384 & 39.36 \\
& C & 384 & 38.86 \\
& S & 384 & 38.90 \\
& T+C & 384 & 40.33 \\
& T+S & 384 & 40.47 \\
& C+S & 384 & 38.49 \\
& T+C+S & 384 & 39.69 \\
\midrule
\multirow{4}{*}{Exact mixture}
& TCS & 384 & 39.69 \\
& DCS & 384 & 40.30 \\
& DAR & 384 & 42.05 \\
& TDC & 384 & 40.56 \\
\bottomrule
\end{tabular}
\captionof{table}{Fixed-budget calibration-set audit on DeepSeek-VL2-Tiny at $\rho=.625$. Avg is the unweighted mean of InfoVQA and MMMU. T/C/S/D/A/R denote TextVQA, ChartQA, ScienceQA, DocVQA, AI2D, and RealWorldQA, respectively.}
\label{tab:calibration_mixture}
\end{minipage}
\par\medskip

%% file: sections/appendix_calibration_route_coverage_table.tex
\par\smallskip
\noindent\begin{minipage}{\columnwidth}
\centering
\small
\setlength{\tabcolsep}{3pt}
\begin{tabular}{@{}lrrr@{}}
\toprule
\textbf{Calibration set} & \textbf{Avg} & \textbf{$J_{16}$} & \textbf{$U_{16}$} \\
\midrule
\rowcolor{rrprow}
TCS & 39.69 & .442 & .411 \\
DCS & 40.30 & .484 & .394 \\
DAR & 42.05 & .477 & .402 \\
TDC & 40.56 & .538 & .378 \\
\bottomrule
\end{tabular}
\captionof{table}{Calibration-set score and routing-coverage diagnostics. $J_{16}$ is the mean top-16 answer-decoding expert Jaccard overlap, and $U_{16}$ is the mean top-16 union coverage.}
\label{tab:calibration_route_coverage}
\end{minipage}
\par\medskip

%% file: sections/appendix_sensitivity_table.tex
\begin{table}[H]
\centering
\small
\setlength{\tabcolsep}{2.2pt}
\begin{tabular}{@{}lcc>{\columncolor{avgcol}}cc@{}}
\toprule
\textbf{Setting} & \textbf{MM3} & \textbf{Text3} & \textbf{Avg} & \textbf{$\Delta$} \\
\midrule
\multicolumn{5}{@{}l}{\textit{Collision weight}} \\
$\lambda=0$ & 50.12 & 58.67 & 54.39 & $-2.55$ \\
$\lambda=.01$ & 53.86 & 59.30 & 56.58 & $-0.36$ \\
\rowcolor{rrprow}
$\lambda=.03$ (ours) & 53.35 & 60.53 & 56.94 & 0.00 \\
$\lambda=.10$ & \textbf{53.93} & \textbf{61.23} & \textbf{57.58} & 0.64 \\
\midrule
\multicolumn{5}{@{}l}{\textit{Grouping rule}} \\
avg. link & 52.94 & 58.83 & 55.89 & $-1.05$ \\
comp. link & 52.27 & 58.33 & 55.30 & $-1.64$ \\
$k$-means++ & $43.7_{\scriptscriptstyle 2.3}$ & $59.2_{\scriptscriptstyle 0.9}$ & $51.5_{\scriptscriptstyle 1.2}$ & $-5.44$ \\
\rowcolor{rrprow}
single link (ours) & \textbf{53.35} & \textbf{60.53} & \textbf{56.94} & 0.00 \\
\bottomrule
\end{tabular}
\caption{DeepSeek-VL2-Tiny Q3 sensitivity at $\rho=.625$ (task-score points; higher is better). $\Delta$ is the Avg change relative to the main setting with $\lambda=.03$ and single linkage. The $k$-means++ row gives five-seed means; subscripts denote standard deviations.}
\label{tab:sensitivity}
\end{table}

%% file: results/aaai27_solver_only_20260717/audit/solver_only_table.tex
\begin{table*}[!t]
\centering
\small
\setlength{\tabcolsep}{3pt}
\begin{tabular}{@{}llrrrrrr>{\columncolor{avgcol}}rr@{}}
\toprule
\textbf{Panel} & \textbf{Method / $\rho$} &
\textbf{Info} & \textbf{MMMU} & \textbf{OCR} &
\textbf{ARC-E} & \textbf{PIQA} & \textbf{Wino} &
\textbf{Avg.} & \textbf{$\Delta J$} \\
\midrule
\rowcolor{rrprow}
A & Greedy/.50 & 49.3 & 32.3 & 70.4 & 49.9 & 65.1 & 52.7 & 53.28 & 0.0\% \\
\rowcolor{rrprow}
A & Greedy/.625 & 50.7 & 35.1 & 74.2 & 58.9 & 69.4 & 53.3 & 56.94 & 0.0\% \\
\rowcolor{rrprow}
A & Greedy/.75 & 59.0 & 37.1 & 77.6 & 62.2 & 70.4 & 56.1 & 60.41 & 0.0\% \\
A & Greedy+1Swap/.50 & 41.4 & 34.2 & 75.1 & 47.1 & 65.1 & 52.9 & 52.6 & -27.35\% \\
A & Greedy+1Swap/.625 & 46.0 & 35.7 & 77.0 & 56.9 & 71.0 & 56.2 & 57.1 & -34.37\% \\
A & Greedy+1Swap/.75 & 47.5 & 38.0 & 79.4 & 61.8 & 72.8 & 58.1 & 59.6 & -42.69\% \\
A & Greedy+SwapC./.50 & 41.1 & 34.7 & 71.8 & 51.1 & 66.2 & 53.7 & 53.1 & -45.60\% \\
A & Greedy+SwapC./.625 & 44.6 & 36.7 & 76.9 & 60.0 & 69.2 & 58.8 & 57.7 & -60.51\% \\
A & Greedy+SwapC./.75 & 47.0 & 37.1 & 79.1 & 62.1 & 72.3 & 59.1 & 59.5 & -73.90\% \\
\midrule
B & avg. link/.625 & 46.2 & 35.2 & 77.4 & 52.4 & 69.4 & 54.7 & 55.9 & -- \\
B & comp. link/.625 & 43.6 & 34.7 & 78.5 & 51.8 & 66.5 & 56.7 & 55.3 & -- \\
B & $k$-means++/.625 &
$34.9\pm3.0$ & $31.0\pm2.4$ & $65.4\pm2.2$ &
$56.3\pm1.8$ & $66.5\pm1.8$ & $54.7\pm1.2$ &
$51.5\pm1.2$ & -- \\
\bottomrule
\end{tabular}
\caption{Controlled search-depth and grouping-rule analysis on DeepSeek-VL2-Tiny using official task scores on a 0--100 scale. Panel A changes only search depth under the same answer-aware merge score and fixed retained expert counts; Panel B changes the grouping rule. Avg is computed before display rounding. The $k$-means++ control reports five seeds as mean$\pm$std.}
\label{tab:solver_only}
\end{table*}

%% file: sections/appendix_small_efficiency_table.tex
\begin{table}[H]
\centering
\footnotesize
\setlength{\tabcolsep}{2.2pt}
\begin{tabular}{@{}lrrr@{}}
\toprule
\textbf{Model} & \textbf{Params (B)} & \textbf{Ckpt. (GiB)} & \textbf{Load (s)} \\
\midrule
\rowcolor{modelrow}
Full & 16.15 & 30.08 & 24.20 \\
\rowcolor{rrprow}
\RoleMerge{} $\rho=.50$ & 8.95 & 16.68 & 12.05 \\
\rowcolor{rrprow}
\RoleMerge{} $\rho=.625$ & 10.75 & 20.03 & 13.02 \\
\rowcolor{rrprow}
\RoleMerge{} $\rho=.75$ & 12.55 & 23.38 & 17.23 \\
\bottomrule
\end{tabular}
\vspace{2pt}

\setlength{\tabcolsep}{1.7pt}
\begin{tabular}{@{}lrrrrr@{}}
\toprule
\textbf{Model} & \textbf{Steady} & \textbf{Peak} & \textbf{TTFT} & \textbf{E2E} & \textbf{tok/s} \\
 & \multicolumn{2}{c}{\textbf{GiB}} & \multicolumn{2}{c}{\textbf{seconds}} & \textbf{32-token} \\
\midrule
\rowcolor{modelrow}
Full & 48.12 & 50.55 & 0.238 & 0.488 & 17.23 \\
\rowcolor{rrprow}
\RoleMerge{} $\rho=.50$ & 31.85 & 34.30 & 0.162 & 0.414 & 18.20 \\
\rowcolor{rrprow}
\RoleMerge{} $\rho=.625$ & 35.92 & 38.37 & 0.183 & 0.459 & 17.91 \\
\rowcolor{rrprow}
\RoleMerge{} $\rho=.75$ & 39.95 & 42.40 & 0.201 & 0.457 & 18.33 \\
\bottomrule
\end{tabular}
\caption{DeepSeek-VL2-Small efficiency on one A100 80GB without CPU/disk offload. Latencies average three run medians over 32 examples per MM3 task (batch size one; 10 warm-ups/task). E2E uses normal generation; tok/s uses 32-token decoding; memory is incremental over pre-load.}
\label{tab:small_single_gpu_efficiency}
\end{table}